\documentclass{article}

\usepackage[final]{neurips_2024}

\usepackage[utf8]{inputenc} 
\usepackage[T1]{fontenc}    
\usepackage{hyperref}       
\usepackage{url}            
\usepackage{booktabs}       
\usepackage{amsfonts}       
\usepackage{nicefrac}       
\usepackage{microtype}      
\usepackage[table]{xcolor}
\usepackage{multirow,multicol}
\usepackage{graphicx}
\usepackage{pifont}
\usepackage{bbding}

\newcommand{\redtext}[1]{\textcolor{red}{#1}}
\newcommand{\greentext}[1]{\textcolor{teal}{#1}}

\usepackage{xspace}
\newcommand{\our}{\textsc{AVerImaTeC}\xspace}

\title{\textsc{AVerImaTeC}: A Dataset for Automatic Verification of Image-Text Claims with Evidence from the Web}

\author{%
  Rui Cao$^\heartsuit$, Zifeng Ding$^\heartsuit$, Zhijiang Guo$^\heartsuit$, Michael Schlichtkrull$^\diamondsuit$, Andreas Vlachos$^\heartsuit$\\
  University of Cambridge$^\heartsuit$, Queen Mary University of London$^\diamondsuit$\\
  \texttt{\{rc990,zd320,zg283,av308\}@cam.ac.uk}, \texttt{m.schlichtkrull@qmul.ac.uk} \\
}

\begin{document}

\maketitle

\begin{abstract}
Textual claims are often accompanied by images to enhance their credibility and spread on social media, but this also raises concerns about the spread of misinformation.
Existing datasets for automated verification of image-text claims remain limited, as  
they often consist of synthetic claims and lack evidence annotations to capture the reasoning behind the verdict.
In this work, we introduce \textbf{\our}, a dataset consisting of 1,297 real-world image-text claims. Each claim is annotated with question-answer (QA) pairs containing evidence from the web, reflecting a decomposed reasoning regarding the verdict.
We mitigate common challenges in fact-checking datasets such as contextual dependence, temporal leakage, and evidence insufficiency, via claim normalization, temporally constrained evidence annotation, and a two-stage sufficiency check. We assess the consistency of the annotation in \our via inter-annotator studies, achieving a $\kappa=0.742$ on verdicts and $74.7\%$ consistency on QA pairs. 
We also propose a novel evaluation method for evidence retrieval and conduct extensive experiments to establish baselines for verifying image-text claims using open-web evidence.
\end{abstract}

\section{Introduction}
\label{sec:intro}
Misinformation has become a public concern due to its potential impact on elections, public health and safety~\citep{99b421ca-b79f-3991-b65a-e1460d47174f,doi:10.1126/science.aaw8243,DBLP:journals/cacm/BarPF23,article-election-pub-health-covid}.
To curb its spread, professional fact-checkers are employed to identify misleading content.
However, they are unable to keep up with the vast volume of information online~\citep{article-fighting-misinfo-scale,10.1093/pnasnexus/pgae217}.
The severity of the problem, along with the limitations of manual verification, has motivated the development of automated fact-checking (AFC)~\citep{DBLP:journals/tacl/GuoSV22,ijcai2021p619}.

To support research in AFC, the research community has created various benchmark datasets~\citep{DBLP:conf/naacl/ThorneVCM18,DBLP:conf/nips/AlyGST00CM21,DBLP:conf/nips/SchlichtkrullG023,DBLP:conf/emnlp/AlhindiPM18,DBLP:conf/sigir/YaoS0CH23,DBLP:conf/naacl/ChenKSDC24}, aiming to enhance the effectiveness and interpretability of fact-checking systems. 
However, most existing benchmarks focus exclusively on textual claims, overlooking the important role of media in the dissemination of misinformation. 
Recent studies estimate that approximately $80\%$ of online claims are multimodal involving both text and media~\citep{DBLP:journals/corr/abs-2405-11697}, as media can enhance perceived credibility~\citep{Newman2012NonprobativeP} and increase exposure~\citep{doi:10.1177/0022243719881113}. 
Among these, images are the most prevalent media type~\citep{DBLP:journals/corr/abs-2405-11697}. 

\begin{figure*}[t] 
	\centering
	\includegraphics[width=0.95\linewidth]{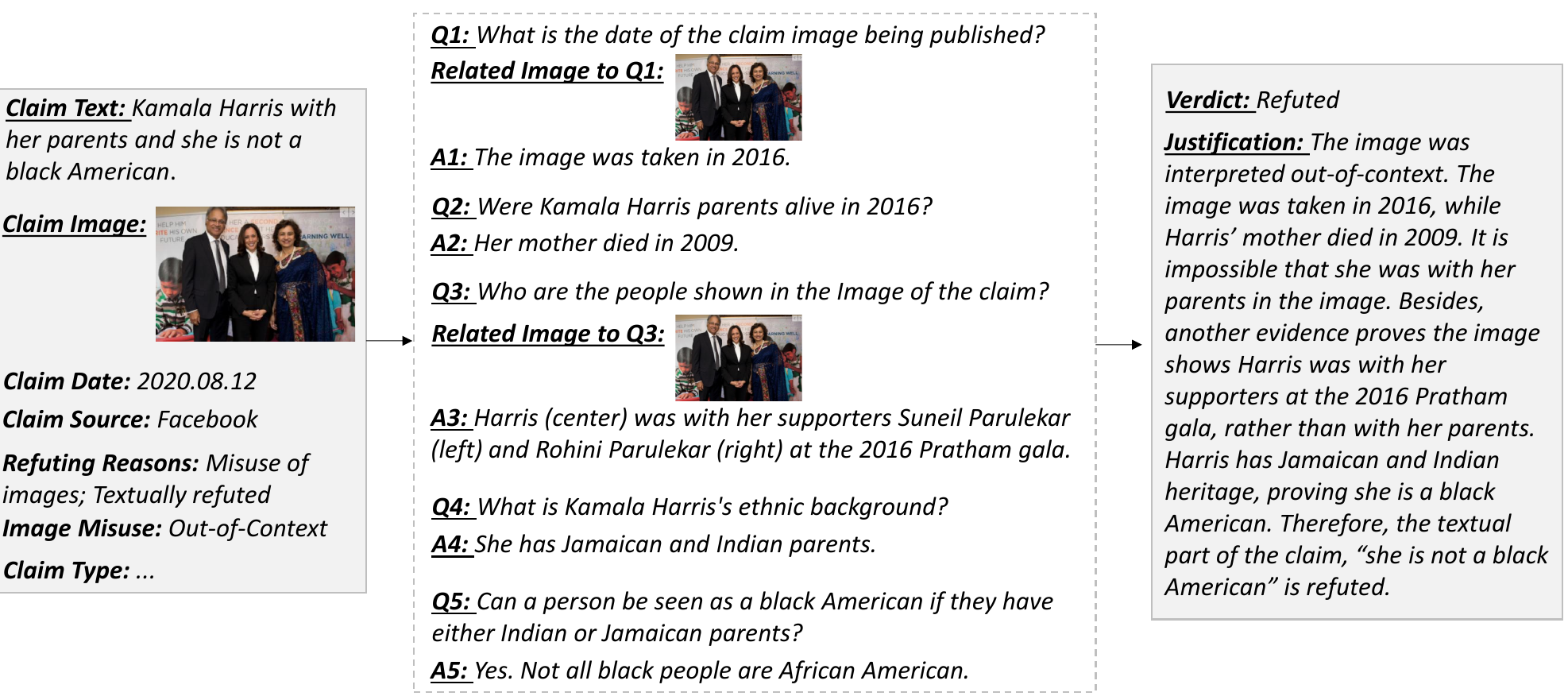} 
	\caption{
\textbf{An annotated claim from \our.} The rationale for verifying an image-text claim has been decomposed into a sequence of QA pairs, which could be potentially multimodal. 
 }
	\label{fig:intro-exp}
\end{figure*}
While several datasets have been developed for image-text AFC, many are synthetic, generated by manually manipulating either the textual or visual modality of image-text pairs~\citep{DBLP:conf/emnlp/LuoDR21,DBLP:journals/ijmir/PapadopoulosKPP24,DBLP:conf/cvpr/JiaHZJCL23}.
Due to discrepancies between synthetic data and real-world data~\citep{DBLP:conf/emnlp/ZengLGP24}, models that perform well on synthetic benchmarks may fail to generalize to real-world claims.
Moreover, recent work~\citep{DBLP:journals/corr/abs-2407-13488} showed that models can achieve high performance on such datasets by exploiting superficial correlations, such as image-text similarity, without examining factuality and logical consistency. 
Some benchmarks~\citep{DBLP:conf/emnlp/ZlatkovaNK19,DBLP:conf/lrec/NakamuraLW20,DBLP:journals/bigdata/ShuMWLL20} attempt to include real-world image-text claims extracted from fact-checking articles. As noted in prior work~\citep{DBLP:conf/emnlp/OusidhoumY022,DBLP:conf/nips/SchlichtkrullG023}, this may result in omitting critical contextual information 
for verification, such as context to resolve coreferences. 
Additionally, both synthetic and real-world image-text datasets typically lack annotated evidence, making it impossible to evaluate models’ reasoning process.

To address the limitations above, we propose the \textbf{A}utomated \textbf{Ver}ification of \textbf{Ima}ge-\textbf{Te}xt \textbf{C}laim (\textbf{\textsc{AVerImaTeC}}) dataset, where the verification of real-world image-text claims has been decomposed into a sequence of question-answering with evidence from the web. 
In addition, each claim is annotated with metadata information, a veracity label based on retrieved evidence and a textual justification, explaining how the verdict is reached, as shown in the example in Figure~\ref{fig:intro-exp}.

To construct \our, 
initially annotators are asked to identify and normalize image-text claims from fact-checking articles, incorporating necessary contextual information while providing associated metadata. 
Next, annotators convert the verification rationale from the articles into QA pairs, while being restricted to using only online evidence published before the claim's date.
Given the multimodal nature of the task, both questions and answers may involve images. Finally, we conduct two rounds of quality control to ensure that each annotated claim is supported by sufficient evidence for the annotated verdict.
The resulting dataset, contains 1,297 image-text claims. 
To assess the consistency of the verdict labels we conducted an inter-annotator agreement study in which we obtained a Randolph's~\citep{randolph} free-marginal $\kappa$ of $0.742$ over 100 re-annotated claims.
The re-annotation
recovered  $74.7\%$ of the original QA pairs, confirming that the annotations capture reasoning paths for verifying image-text claims consistently.

We further introduce a baseline for image-text claim verification, which operates by generating evidence-seeking questions aimed at fact-checking and answering these questions with a set of expert tools. 
Since \our is the first image-text claim verification dataset to incorporate QA annotations that explicitly reflect reasoning paths and evidence, we develop a reference-based evaluation method to assess models' generated questions and retrieved evidence. Using this evidence evaluation, we report conditional verdict accuracy which measures the correctness of predicted verdicts only when the associated evidence score exceeds a predefined threshold\footnote{The dataset is available here: https://huggingface.co/datasets/Rui4416/AVerImaTeC. The code can be accessed here: https://github.com/abril4416/AVerImaTeC}.

\section{Related Works}
\label{sec:related}

\begin{table*}[t]
\centering
\small
\caption{\textbf{Comparison of fact-checking datasets.} 
\textit{Indep.} (independence) denotes whether extracted claims are context independent (e.g., understandable without fact-checking articles). \textit{Img.}, \textit{Suff.}, \textit{Retr.}, and \textit{Unleak.} are abbreviations for image, sufficiency, retrieval and unleaked evidence, respectively.
Sufficiency indicates whether there is sufficient supporting evidence to reach annotated verdicts; Retrieval refers to whether open-world evidence retrieval is performed; Unleaked represents whether annotated evidence contains temporal leakage such as including evidence published after claims.
}
  \label{tab:rela-dataset}
  \begin{tabular}{lccccccc}
    \toprule
    \multirow{2.5}{*}{\textbf{Dataset}}&
    \multicolumn{2}{c}{\textbf{Claim}} & \multicolumn{4}{c}{\textbf{Evidence}} & \multirow{2.5}{*}{\textbf{\# Claims}} \\
   \cmidrule(lr){2-3} \cmidrule(lr){4-7}
   & \textit{Real} & \textit{Indep.} &
   \textit{Img.} &
     \textit{Suff.} & \textit{Retr.} & \textit{Unleak.} &\\
    \midrule
    FEVER~\citep{DBLP:conf/naacl/ThorneVCM18}  & \redtext{\ding{56}}  &\greentext{\checkmark} &\redtext{\ding{56}} & \greentext{\checkmark}&\greentext{\checkmark} &-  &185,445\\
    FEVEROUS~\citep{DBLP:conf/nips/AlyGST00CM21}   & \redtext{\ding{56}}&\greentext{\checkmark} &\redtext{\ding{56}}&\greentext{\checkmark} &\greentext{\checkmark} &  -&87,026\\
    Liar-Plus~\citep{DBLP:conf/emnlp/AlhindiPM18}   &\greentext{\checkmark} &\redtext{\ding{56}} &\redtext{\ding{56}} &\greentext{\checkmark} &\redtext{\ding{56}} & \redtext{\ding{56}} &12,836\\
    Snopes~\citep{DBLP:conf/conll/HanselowskiSSLG19}   &\greentext{\checkmark} &\redtext{\ding{56}} & \redtext{\ding{56}}& \redtext{\ding{56}}&\redtext{\ding{56}} & \greentext{\checkmark} &6,422\\
    MultiFC~\citep{DBLP:conf/emnlp/AugensteinLWLHH19}   &\greentext{\checkmark} &\redtext{\ding{56}} & \redtext{\ding{56}}&\redtext{\ding{56}} &\greentext{\checkmark} & \redtext{\ding{56}} &36,534\\
    AVeriTec~\citep{DBLP:conf/nips/SchlichtkrullG023}  &\greentext{\checkmark} &\greentext{\checkmark} & \redtext{\ding{56}}&\greentext{\checkmark} &\greentext{\checkmark} & \greentext{\checkmark} &4,568\\
    CLAIMDECOMP~\citep{DBLP:conf/naacl/ChenKSDC24}    & \greentext{\checkmark} & \redtext{\ding{56}} & \redtext{\ding{56}}&\redtext{\ding{56}}  &\greentext{\checkmark} & \greentext{\checkmark} &1,200\\
    MOCHEG~\citep{DBLP:conf/sigir/YaoS0CH23}  &\greentext{\checkmark} & \redtext{\ding{56}}&\greentext{\checkmark} &\redtext{\ding{56}} & \redtext{\ding{56}}& \redtext{\ding{56}} &15,601\\
    \midrule
    NewsCLIPpings~\citep{DBLP:conf/emnlp/LuoDR21}  &\redtext{\ding{56}} &\greentext{\checkmark} & -&- & -& - &988,283\\
    InfoSurgeon~\citep{DBLP:conf/acl/FungTRPJCMBS20} &\redtext{\ding{56}} &\greentext{\checkmark} & -&- & -& - &30,000\\
    Autosplice~\citep{DBLP:conf/cvpr/JiaHZJCL23}  &\redtext{\ding{56}} &\greentext{\checkmark} & -&- & -& - &5,894\\
    DGM~\citep{DBLP:conf/cvpr/Shao0L23}  & \redtext{\ding{56}}& \greentext{\checkmark}& -&- & -& - &230,000\\
    MMFake~\citep{DBLP:journals/corr/abs-2406-08772} &\redtext{\ding{56}}  &\greentext{\checkmark} & -&- & -& - &11,000\\
    Verite~\citep{DBLP:journals/ijmir/PapadopoulosKPP24} &\redtext{\ding{56}} &\greentext{\checkmark} & -&- & -& - &1,000\\
    COSMOS~\citep{DBLP:journals/corr/abs-2101-06278} &Mix &\redtext{\ding{56}} & -&- & -& - &201,700\\
    FACTIFY~\citep{DBLP:conf/aaai/MishraSBCRPD0SE22}  & Mix & \redtext{\ding{56}}& -&- &- & - &50,000\\
    FACTIFY 2~\citep{DBLP:conf/defactify/SuryavardanMPCR23}  & Mix  & \redtext{\ding{56}}& -&- &- & - &50,000\\
    MMOOC~\citep{DBLP:conf/acisp/XuDCLY24}  & Mix & \greentext{\checkmark}& -&- &- & - &364,000\\
    Fauxtography~\citep{DBLP:conf/emnlp/ZlatkovaNK19}  & \greentext{\checkmark}&\redtext{\ding{56}} &- &- &- &- &1,233\\
    Fakeddit~\citep{DBLP:conf/lrec/NakamuraLW20}  &\greentext{\checkmark}  & \redtext{\ding{56}}& -&- & -& - &1,063,106\\
    Qprop~\citep{DBLP:journals/ipm/Barron-CedenoJM19}  & \greentext{\checkmark}& \redtext{\ding{56}} & -&- & -& - &51,294\\
    FakeNewsNet~\citep{DBLP:journals/bigdata/ShuMWLL20}  &  \greentext{\checkmark}&\redtext{\ding{56}}  & -&- & -& - &23,196\\
    MuMiN~\citep{DBLP:conf/sigir/NielsenM22}  & \greentext{\checkmark}&\redtext{\ding{56}} & -&- & -& - &12,914\\
    \midrule
    \our& \greentext{\checkmark}&\greentext{\checkmark}&\greentext{\checkmark}&\greentext{\checkmark}&\greentext{\checkmark}&\greentext{\checkmark} &1,297\\
    \bottomrule
\end{tabular}
\end{table*}

Automated fact-checking (AFC) has become increasingly important due to the pressing need to curb the spread of misinformation~\citep{DBLP:journals/tacl/GuoSV22,ijcai2021p619}. To support AFC research, several  datasets focusing primarily on text-based claims have been proposed (see the top block of Table~\ref{tab:rela-dataset}). 
Motivated by the prevalence of images in claims, fact-checking datasets for image-text claims were introduced (listed in the bottom block of Table~\ref{tab:rela-dataset}). Some studies~\citep{DBLP:conf/emnlp/LuoDR21,DBLP:journals/ijmir/PapadopoulosKPP24,DBLP:conf/cvpr/JiaHZJCL23} have generated synthetic claims by applying manipulation techniques to the visual and textual modalities of image-text pairs. However, there are discrepancies between synthetic data and real-world image-text claims~\citep{DBLP:conf/emnlp/ZengLGP24,DBLP:journals/corr/abs-2407-13488}, raising concerns about the generalization of models to real-world image-text claim verification.
Although some benchmarks~\citep{DBLP:conf/emnlp/ZlatkovaNK19,DBLP:journals/ipm/Barron-CedenoJM19,DBLP:journals/bigdata/ShuMWLL20} focus on real-world claims (e.g., those derived from fact-checking articles), they all suffer from context dependence. Moreover, all existing datasets with image-text claims lack evidence annotations, limiting transparency,  and the ability to understand the rationale behind fact-checking verdicts.

To capture the rationale in claim verification, a complex reasoning task, various reasoning representations have been explored, including natural logic~\citep{DBLP:conf/emnlp/StrongA024}, functional programs~\citep{DBLP:conf/acl/PanWLLWKN23}, and QA~\citep{DBLP:conf/naacl/ChenKSDC24,DBLP:conf/emnlp/QiXSLJWH23,DBLP:conf/emnlp/PanLKN23}.
In AVerImaTeC, we adopt QA as the reasoning representation, as the QA format is more intuitive and accessible for annotators, enabling efficient and consistent annotation by non-experts. Furthermore, other reasoning forms can often be mapped into QA-style representations, ensuring compatibility and flexibility for future extensions.

\section{Annotation Schema}
\label{sec:anno_struct}
Each claim in \our is normalised to be understandable alone without additional context, such as the original social media post or the associated fact-checking article.
For each claim, we provide key metadata such as the \textit{speaker}, \textit{publisher}, \textit{publication date}, and the relevant \textit{location}.
These metadata elements can serve as valuable evidence for claim verification. 
We further annotate each claim with 
\textit{claim type} (e.g., \textit{quote verification}, which determines whether a quote was actually attributed to the correct speaker), and \textit{fact-checking strategy} (e.g., \textit{reverse image search} to find background information about images).
While not all of these annotations are directly used during verification, they offer insights for developing fact-checking models. 

The sequence of QA pairs reflects the reasoning process involved in evidence retrieval and verification. 
Given the multimodal nature of the claims, both questions and answers may include images (i.e., image-related questions and image answers).
A single question may have multiple answers,
as there may be conflicts or disagreements in the evidence. 
Questions may refer to previous questions and their answers as long as they are understandable on their own, thus capturing multi-hop reasoning in fact-checking. 
Answers (other than \textit{``No answer could be found.''} or derived via image analysis without external information) must be supported by a \textit{source url} linking to a web page. 
To ensure long-term accessibility, all source pages are archived on the internet archive.\footnote{https://archive.org/}
We also provide metadata for each annotated QA pair, including the \textit{question type} (e.g.,  \textit{image-related} or \textit{metadata-related}), \textit{answering method}, \textit{answer type} (e.g., \textit{extractive}, \textit{abstractive}, \textit{boolean} or \textit{image-based}), and \textit{source medium type} (type of web content used as evidence).

We follow the four-way veracity labeling schema from~\cite{DBLP:conf/nips/SchlichtkrullG023}: \textit{supported}, \textit{refuted}, \textit{not enough evidence}, and \textit{conflicting/cherry-picking}.
An image-text claim can be refuted due to its textual part (the textual part of is factually wrong) 
and/or due to \textit{misuse of images} (e.g., mis-interpreting the context of an image). 
For instance, the claim in Figure~\ref{fig:intro-exp} is refuted due to both textual refutation and image misuse. 
\textit{Not enough evidence} refers to cases where evidence is insufficient to either support or refute a claim. \textit{Conflicting/Cherry-picking} covers cherry-picking claims,
true-but-misleading claims, as well as claims with conflicting evidence. Conflicts among evidence has been extensively studied in the context of QA~\citep{DBLP:journals/corr/abs-2410-12311,DBLP:journals/corr/abs-2404-12447}. 

A textual justification is added to the claims to explain how verdicts can be reached on the basis of the evidence found. Considering that the evidence can contain multiple images, we assign unique and special tokens to the images in the evidence (e.g., \texttt{[IMG\_1]}) for annotators to refer to in justifications. 
They may also include commonsense reasoning or inductive reasoning beyond the retrieved evidence. For instance, the evidence of the claim in Figure~\ref{fig:intro-exp} proves the image was taken in 2016 while Kamala Harris' mother died in 2009. The justification infers it is impossible that Harris \textit{was with her parents in
the image}, taken in 2016 as her mother died in 2009 already.

\section{Annotation Process}
\label{sec:anno_pipe}
\begin{figure*}[t] 
	\centering
	\includegraphics[width=\linewidth]{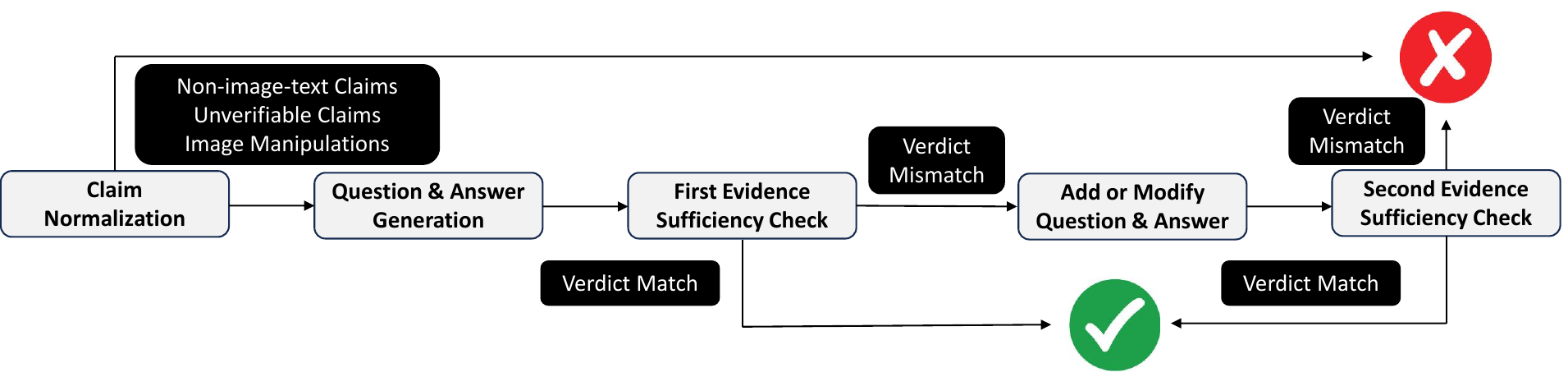} 
	\caption{
 \textbf{Annotation pipeline.} We first 
 normalize the claim, then perform QA annotation to structure evidence retrieval. 
 Two rounds of evidence sufficiency checks ensure annotation quality.} 
	\label{fig:anno-pipeline}
\end{figure*}

The five-phase annotation pipeline is illustrated in Figure~\ref{fig:anno-pipeline}, extending the annotation process proposed in~\cite{DBLP:conf/nips/SchlichtkrullG023} to the domain of image-text claims.
In phase one, an annotator extracts and normalize valid image-text claims, given a fact-checking article. For each extracted claim, a different annotator in the second phase generates questions and answers to reflect the rationale of fact-checking based on the article, using evidence from the web. A provisional veracity label is annotated as well. Thirdly, a first round of evidence sufficiency checking will be conducted by a third annotator who provides a justification and a verdict solely based on the QA pairs, without considering the fact-checking article.
Different verdicts in the second and third phases suggest possible insufficient evidence in the second phase. In this case, the claim is forwarded to a fourth phase to add or modify existing QA pairs and
a fifth phase for an additional round of evidence sufficiency check. 
Any claims for with unresolved conflicts in the verdicts are discarded.
We ensure that each annotation phase of a claim is done independently 
by using different annotators (i.e., the same claim will not be annotated by the same annotator twice).
Details for annotation guidelines and annotators' demographics are provided in Appendix~\ref{sec:app-annotation-guidelines} and~\ref{sec:app-anno-detail-annotator-demo}, respectively. 

\noindent\textbf{Claim Extraction \& Normalization. }Given a fact-checking article, an annotator extracts all claims from it, as multiple claims may pertain to a single event. 
The modality types of claims are annotated, and only \textit{image-text} claims (i.e., textual claims paired with images) are forwarded to phase two.
For identified image-text claims, sufficient context must be provided to ensure that the claim, specifically its textual component, can be understood  independently of the surrounding article. 
In some cases, metadata (e.g., date, location, speaker) is sufficient to disambiguate a claim. However, for ambiguities beyond metadata, such as unresolved coreference or missing referents, annotators are instructed to enrich the extracted claims so that they can be interpreted independently. 
The associated images are uploaded and normalized (e.g., separating collages or locating original, unaltered versions).
Relevant metadata is also annotated.
We exclude unverifiable claims (e.g.,  speculation or personal opinions) 
and  claims where images are not used in verification or involve manipulated content.

\noindent\textbf{Question Generation \& Answering. }Annotators in this phase are instructed to transform the rationales derived from fact-checking articles into a sequence of QA pairs. 
Each question is annotated with its \textit{question type}, \textit{answering method} and \textit{answers}. 
For \textit{image-related} questions, annotators must select relevant images from either the claim images or images from previous answers. If a question is not marked as \textit{unanswerable} or pertains to \textit{image analysis}, annotators are required to provide supporting urls of the evidence source.
We advise annotators to prioritize evidence sources linked within the fact-checking articles but to exclude anything published after the claim date, including the article itself.
When linked sources are unavailable (e.g., dead links) or insufficient, annotators are provided with a custom Google search interface that supports both text and image queries.
All retrieved pages from the interface are restricted to dates \textit{prior} to the claim date to prevent temporal leakage~\citep{DBLP:conf/emnlp/Glockner0G22}. Based on the generated QA pairs, annotators assign a verdict.

\noindent\textbf{Evidence Sufficiency Check. }
In this phase, 
a third annotator, who does not have access to the fact-checking article, is presented with the extracted claim and its associated annotated QA pairs.
The annotator is tasked with assigning a verdict and providing a textual justification. 
This verdict is then compared to the one generated during the QA annotation phase. A discrepancy between the two verdicts indicates insufficient evidence, and the QA annotation process is repeated to refine the QA pairs, followed by a second round of evidence sufficiency assessment with new annotators.
\section{Dataset Statistics}
\label{sec:data_stat}
\begin{table*}[t]
\small
\centering
\caption{\textbf{Data statistics for dataset splits.} End date refers to the latest publication date of claims included in each split. The start date of each \textit{dev} and \textit{test} split corresponds to the end date of the preceding split. The final row of the table reports the distribution of claim labels across four categories: \textit{supported} (S), \textit{refuted} (R), \textit{conflicting/cherry-picking} (C), and \textit{not enough evidence} (N).}
  \label{tab:data-statistics}
  \begin{tabular}{l|ccc}
    \toprule
    \textbf{Split} &\textbf{Train} &\textbf{Dev} &\textbf{Test} \\
    \midrule
    \# Claims & 793&152 &352 \\
    \# Images / Claim  & 1.49&1.38 & 1.38\\
    \# QA Pairs / Claim & 2.86& 2.84& 3.11\\
    Reannotated (\%) &15.0 &15.8 &9.4 \\
    End Date &31-05-2023 &31-07-2023 &21-03-2025 \\
    Labels (S / R / C / N) (\%) & 1.6 / 95.3 / 0.8 / 2.3& 2.6 / 92.8 / 0.7 / 3.9& 13.9 / 78.1 / 2.0 / 6.0\\
    \bottomrule
\end{tabular}
\end{table*}

\noindent\textbf{Data Distribution. } We began with 2,353 fact-checking articles
After discarding those that were inaccessible, did not focus on image-text claims, or contained unresolved annotation conflicts, we obtained 1,297 annotated image-text claims using the annotation pipeline described in the previous section.
Further details on article sources and the filtering process are provided in Appendix~\ref{sec:app-data-p1-data-filter} and~\ref{sec:app-discard-claims}, respectively.
The splits of our dataset are temporally organized, and detailed statistics are presented in Table~\ref{tab:data-statistics}. 
We find that $23.7\%$ image-text claims include more than one claim image,  highlighting the need to understand multiple visual inputs.
On average, each claim is annotated with $2.92$ questions. Of these, $3.5\%$ questions have more than one answer, and 
$62.5\%$ are image-related, emphasizing the importance of visual context in verifying image-text claims. 
To address these questions, annotators frequently selected
\textit{Image-search} ($53.9\%$) as the answering method, indicating the necessity for tools that support image-centric information retrieval.
Regarding answer types, 
$58.8\%$ are \textit{extractive}, consistent with our annotation guidelines. Additionally, $1.6\%$ of answers are images themselves, underscoring the importance of supporting image retrieval as direct answers.
A small proportion ($2.6\%$) of questions are marked as \textit{unanswerable}, reflecting cases where no supporting evidence could be found online.
Further metadata statistics, such as \textit{claim types} and \textit{answer types}, are provided in Appendix~\ref{sec:app-data-statistics-metadatao}.
The dataset shows a label imbalance, with most claims being \textit{refuted}, which is expected given that  misleading content is more likely to be scrutinized.

\noindent\textbf{Inter-Annotator Agreement. }
We re-annotated $100$ claims with a different group of annotators, following the same annotation pipeline as described in Section~\ref{sec:anno_pipe}. As in prior work~\citep{DBLP:conf/nips/SchlichtkrullG023}, we assume that 
the first phase of annotation
has already been completed, and thus the re-annotation process begins from the second phase. We evaluate inter-annotator agreement for both verdict labels and QA annotations.
For verdict agreement, we use
Randolph's~\citep{randolph} free-marginal multi-rater $\kappa$, designed for unbalanced datasets~\citep{10.1007/s11634-010-0073-4}. We obtained an agreement score of $\kappa = \mathbf{0.742}$. For comparison, 
AVeriTec~\citep{DBLP:conf/nips/SchlichtkrullG023} reported agreement of $0.619$. 
Using Fleiss’ $\kappa$, a more traditional metric, our annotation process achieves an agreement score of $0.450$.
To evaluate QA annotations, the three best performing annotators were provided with an extracted claim and two independently annotated sets of QA pairs, and asked to determine how many QA pairs in one set were covered by the other.
We compute recall and precision by comparing the original annotated QA pairs against those from re-annotation. The recall rate is $\mathbf{74.7\%}$ and the precision is $\mathbf{67.2\%}$. 
The substantial overlap between the two sets suggests strong agreement between annotators. 
\section{Evaluation}
\label{sec:eval}
The evaluation of model accuracy on our dataset considers both the \textit{retrieved evidence} and the \textit{veracity}.
Following~\citep{DBLP:conf/naacl/ThorneVCM18,DBLP:conf/nips/SchlichtkrullG023}, we first assess the quality of the retrieved evidence by comparing it against human-annotated references, and report veracity prediction accuracy conditioned on the evidence scores. I.e., the accuracy of a verdict prediction is considered only if the associated evidence score exceeds a predefined threshold $\lambda$, otherwise the claim is considered to be labeled incorrectly.
This reflects the requirement that an effective fact-checking system should not only predict return verdicts on claims but also provide appropriate evidence.

Recent research~\citep{DBLP:journals/corr/abs-2411-05375} showed that reference-based evaluation of evidence with large language models (LLMs) aligns best with human assessments. 
We extend this framework to a multimodal setting where evidence from the web may comprise text and images. We transform each QA pair into an evidence statement following~\citep{DBLP:journals/corr/abs-2411-05375}, where images are represented with special image tokens (e.g., \texttt{[IMG\_1]}). For instance, the first QA pair in Figure~\ref{fig:intro-exp} is transformed to \textit{``[IMG\_1] was published in 2016.''}.
We then conduct separate reference-based evaluations for the textual and visual components. For the textual part, we use a validation method similar to~\citep{DBLP:journals/corr/abs-2411-05375}.
If a matched evidence item is found within the ground-truth set, we proceed to a second step to compare the associated images. If image similarity falls below a threshold 
the match is deemed invalid due to image mismatch. 
We exploit Gemini-2.0-Flash~\citep{gemini-deepmind} as the scoring model for both steps in the reference-based evaluation inspired by its power~\citep{DBLP:journals/corr/abs-2411-05375}.
We report evidence \textit{recall}, defined as the percentage of ground-truth evidence instances successfully retrieved.

We performed alignment checks and robustness checks (against adversarial attacks) of different reference-based evaluation schemes (text-only, interleaved and the separated evaluation) in our setting (details in Appendix~\ref{sec:app-eval-evid-eval} and~\ref{sec:app-eval-robust}). Our separated reference-based evaluation achieved the highest alignment with human assessments, with a Spearman correlation coefficient ($\rho$)~\citep{spearman04} of $0.332$ and a Pearson correlation coefficient ($r$)~\citep{pearson1896mathematical} of $0.381$. Furthermore, the separated evaluation method is the most robust towards adversarial attacks.

Our evaluation primarily focuses on evidence retrieval and verdict prediction. 
Additionally, given the availability of justification annotations, we assess the quality of model-generated justifications by comparing them to the human-annotated ground truth with the traditional evaluation metric, ROUGE-1~\citep{lin-2004-rouge}, a standard metric that is relatively tolerant of longer outputs.
While ROUGE-1 provides a baseline assessment, we recognize that more sophisticated and targeted evaluation methods may be necessary. 
Similarly, we adopt a conditional justification generation score that any claim for which the
evidence score is below $\lambda$ receives a score for justification of 0.
The prompts used for the QA pair conversion and evaluation are provided in Appendix~\ref{sec:app-prompt-qa-to-evid} and~\ref{sec:app-prompt-evaluation}, respectively.

\section{Experiments}
\label{sec:experiment}

\subsection{Baselines}
\label{sec:exp-baseline}
Our baseline system guides the fact-checking process by sequentially posing and answering essential questions to verify image-text claims. It consists of four  components: 
a \textit{question generator}, an \textit{answer generator}, a \textit{verifier} and a \textit{justification generator}.
Throughout the question-answering process, the system maintains an evolving \textit{evidence history context}, which is continuously updated with summarized evidence derived from each QA pair.
After the QA stage, the verifier receives the claim and the accumulated evidence context to predict a veracity label. 
Finally, the justification generator produces a rationale that explains how the predicted verdict is supported by the evidence.

\noindent\textbf{Question Generator.} A straightforward approach is to generate all verification questions at once, given an image-text claim, as in prior work~\citep{DBLP:conf/nips/SchlichtkrullG023}. We refer to this strategy as \textit{paralleled} question generation (\textbf{PQG}). 
However, since later questions often depend on earlier ones (e.g., as shown in Figure~\ref{fig:intro-exp}), we also propose a \textit{dynamic} question generation (\textbf{DQG}) method, where each subsequent question is generated based on both the claim and the evolving evidence history.
To combine the strengths of both approaches, we introduce a \textit{hybrid} question generation (\textbf{HQA}) method, which generates the first few questions in parallel and then switches to dynamic generation for the remaining ones. 
All three strategies employ an MLLM for question generation, leveraging the model's internal decomposition capability.
The prompts used for each strategy are detailed in Appendix~\ref{sec:app-prompt-qg}.
\begin{table*}[t]
\centering
\small
\caption{\textbf{Experimental results of baselines on \our.} \textbf{Q-Eval} and \textbf{Evid-Eval} denote for \textit{recall} scores of generated questions and retrieved evidence, with reference of ground-truth questions and evidence. We report verdict prediction and justification generation scores conditioned on evidence retrieval performance, specifically only considering verdict accuracy and justification generation performance when the evidence score is above 0.2, 0.3 and 0.4.}
  \label{tab:exp-results-main}
  \begin{tabular}{ll|cc|ccc|ccc}
    \toprule
    \textbf{LLM} &\textbf{MLLM} &\textbf{Q-Eval} &\textbf{Evid-Eval} & \multicolumn{3}{c}{\textbf{Veracity} \@ (.2/.3/.4)} & \multicolumn{3}{c}{\textbf{Justifications} \@ (.2/.3/.4)}\\
    \midrule
    \rowcolor{lightgray}
   \multicolumn{10}{c}{\textit{Paralleled Question Generation}}\\ 
    \midrule
    Gemini & Gemini &0.42 & 0.15&0.15 &0.13 & 0.08& 0.15&0.11 &0.07 \\
    Qwen & Qwen-VL &0.43 &0.18 &0.09& 0.08&0.05 &0.13 &0.11 &0.07 \\
    Gemma & Gemma  & 0.39& 0.21&0.14 & 0.12&0.09 &0.17 & 0.14& 0.10\\
    Qwen & LLaVA  & 0.37& 0.16& 0.09&0.08 &0.05 &0.12 & 0.10&0.06 \\
    \midrule
    \rowcolor{lightgray}
   \multicolumn{10}{c}{\textit{Dynamic Question Generation}}\\ 
    \midrule
     Gemini & Gemini &0.33& 0.22&0.17 &0.16 &0.12 &0.17 &0.16&0.11\\
    Qwen & Qwen-VL &0.27 & 0.12&0.10 & 0.09&0.05 &0.09 &0.08 &0.05\\
    Gemma & Gemma  & 0.27& 0.19& 0.15&0.13 &0.10 & 0.15& 0.13&0.09 \\
    Qwen & LLaVA  & 0.32&0.16 & 0.13& 0.11& 0.08&0.11 & 0.10&0.08 \\
    \midrule
    \rowcolor{lightgray}
   \multicolumn{10}{c}{\textit{Hybrid Question Generation}}\\ 
    \midrule
     Gemini & Gemini & 0.36&0.19 &0.18 &0.17 & 0.10&0.17 &0.15& 0.09\\
    Qwen & Qwen-VL &0.37 &0.16 &0.11 &0.09 & 0.06& 0.12& 0.10&0.06 \\
    Gemma & Gemma  &0.26 &0.25 &0.16 &0.15 &0.11 &0.19 &0.17& 0.12\\
    Qwen & LLaVA  & 0.30&0.17 &0.09 &0.09 & 0.07&0.12 &0.11&0.07 \\
    \bottomrule
\end{tabular}
\end{table*}

\noindent\textbf{Answer Generator. }
Given a generated question, the answer generator is responsible for generating an answer to the question.
Inspired by recent research on tool usage~\citep{DBLP:journals/corr/abs-2303-04671,DBLP:journals/corr/abs-2411-13697,DBLP:journals/corr/abs-2412-10510}, we integrate a set of specialized tools into the answer generation module, along with a tool selector that automatically selects the appropriate tool for a given question.
The system includes tools for: (1) \textit{reverse image search} (RIS) to retrieve text-based information associated with an image; (2) \textit{web search for texts} (WST) to retrieve relevant web texts given a textual query; (3) \textit{web search for images} (WSI) to retrieve relevant images using a textual query; and (4) \textit{visual question answering} (VQA) to answer questions directly based on input images, including comparison and detail analysis.
Among these, VQA, implemented with an MLLM, directly outputs an answer, while RIS and WST are followed by an LLM to leverage the retrieved text to generate an answer.
WSI is followed by an MLLM, which incorporates the retrieved image into the answer generation process. 
Prompts for tool selection and answer generation are provided in Appendix~\ref{sec:app-prompt-tool-sel} and~\ref{sec:app-prompt-ans-gen}, respectively. 

\noindent\textbf{Verifier. }The verifier takes the claim to be verified, the multimodal evidence context and the verdict definitions as introduced in Section~\ref{sec:anno_struct} to predict a veracity label for the claim
(prompts in use shown in Appendix~\ref{sec:app-prompt-verdict-gen}). An MLLM will serve as the verifier.

\noindent\textbf{Justification Generator. }Given the predicted verdict, in this step, the justification generator is asked to provide an explanation for the prediction. It also has access to the claim and evidence history (detailed prompts in Appendix~\ref{sec:app-prompt-justi-gen}).
We rely on an MLLM for justification generation.

\noindent\textbf{Model Implementation. }
Our baseline system includes both an LLM and an MLLM, which could take different roles in components. 
We experimented with four combinations of LLMs and MLLMs: 1) Gemini-2.0-flash-001~\citep{gemini-deepmind} (\textbf{Gemini}) as both the LLM and the MLLM; 2) Qwen/Qwen2.5-7B-Instruct~\citep{DBLP:journals/corr/abs-2412-15115} (\textbf{Qwen}) acts as the LLM and Qwen2.5-VL-7B-Instruct~\citep{DBLP:journals/corr/abs-2502-13923} (\textbf{Qwen-VL}) serves as the MLLM; 3) Gemma-3-12B~\citep{DBLP:journals/corr/abs-2503-19786} (\textbf{Gemma}), capable of both unimodal and multimodal understanding, is used as both the LLM and the MLLM; and 4) Qwen and LLaVA-Next-7B~\citep{DBLP:journals/corr/abs-2407-07895} (\textbf{LLaVA}) work as the LLM and MLLM respectively. 
More details on model implementation and experiment settings are  in Appendix~\ref{sec:app-detail-of-exp}.

\begin{table*}[t]
\centering
\small
\caption{\textbf{Baseline performance on verdict prediction and justification generation with ground-truth evidence (first block) and without accessing any external evidence (second block).}
NEE is for \textit{Not Enough Evidence} and Conflict.\ is for \textit{Conflicting/Cherry-picking}.
Justi.\ is for the performance of justification generation.}
  \label{tab:exp-ablation-results}
  \begin{tabular}{l|ll|cccc|c|c}
    \toprule
   \textbf{Evid. Source} & \textbf{LLM} &\textbf{MLLM} &\textbf{Refuted} &\textbf{Supported} & \textbf{NEE} &\textbf{Conflict.} &\textbf{Overall}& \textbf{Justi.}\\
    \midrule
   \multirow{4}{*}{Ground-Truth}& Gemini & Gemini & 0.84&0.92 &0.52& 0.00&0.82&0.50\\
   & Qwen & Qwen-VL &0.87 & 0.47&0.62 & 0.00&0.78&0.44\\
   & Gemma & Gemma &0.63 &0.84 & 0.62& 0.00&0.64& 0.49\\
   & Qwen & LLaVA &0.55 & 0.47&0.90 &0.00 &0.55 & 0.43\\
    \midrule
   No Search & Qwen &Qwen-VL  &0.01 &0.02 &0.14 & 0.00&0.02 & 0.04 \\
    \bottomrule
\end{tabular}
\end{table*}

\subsection{Main Results}
\label{sec:exp-results}
We investigated the performance of baseline models both under zero-shot and few-shot settings.  
In the latter, three training instances were used as demonstrations to guide question generation (details in Appendix~\ref{sec:app-detail-exp-setting}). 
Overall, few-shot baselines slightly outperform their zero-shot counterparts. 
Few-shot performance results for various combinations of LLMs and MLLMs are presented in Table~\ref{tab:exp-results-main} (zero-shot model performance in Appendix~\ref{sec:app-add-exp-zero-shot}, and the findings hold for both settings). 
We report conditional veracity accuracy and ROUGE-1 scores for justification generation under varying evidence evaluation thresholds, $\lambda = \{0.2, 0.3, 0.4\}$.

\noindent\textbf{Comparison of Question Generation Strategies. }
Among the three question generation strategies, the parallel approach consistently outperformed others in producing critical questions for fact-checking. 
The dynamic strategy, where questions are generated based on evolving evidence, yielded a weaker performance. 
This suggests that the increased complexity of reasoning over evolving interleaved image-text evidence poses significant challenges for current MLLMs.
Open-source MLLMs often generated repetitive questions when using the dynamic strategy, further highlighting their limitations in handling complex multimodal inputs. 

\textbf{Performance of Evidence Retrieval. }
Evidence scores across all models are significantly lower than their question evaluation scores, underscoring the difficulty of retrieving appropriate evidence for image-text verification.
One reason for the failure of evidence retrieval is that models exhibited a bias toward using VQA as the answering tool (e.g., Qwen + Qwen-VL in PQG selected VQA as the answering tool for $30\%$ questions), diverging from human fact-checkers' preferences, despite being provided with tool selection demonstration examples
(more elaborations in Appendix~\ref{sec:app-detail-of-exp}).
MLLMs tend to rely more on internal image details rather than external contextual information to address image-related questions.
Additionally, approximately $13\%$ images failed to retrieve any contextual information via RIS (i.e., no web pages published before claim dates could be found), consistent with the findings of~\citep{DBLP:conf/emnlp/TongletMG24}. 
A notable portion of questions, e.g., $30\%$ questions of the Qwen + Qwen-VL baseline with PQG, elicited responses such as \textit{``No answer could be found''}, based on retrieved evidence. 
This can be attributed to two main factors besides the failure of RIS: (1) many web pages retrieved by RIS were non-scrapable (e.g., Instagram posts), and (2) the baseline evidence employed a naive ranking method, BM25~\citep{DBLP:journals/ftir/RobertsonZ09}, which neither considered the visual content of images nor incorporated fine-grained re-ranking.
Interestingly, higher scores for question generation did not always translate into better evidence retrieval. 
For instance, under the hybrid strategy, the Gemini-based baseline model achieved a 0.1-point higher question generation score than the Gemma-based model, but had worse evidence retrieval performance.
Further analysis showed that Gemini generated a higher proportion of RIS-dependent questions ($40.1\%$ vs. $32.4\%$), and the majority ($62.6\%$ vs. $31.6\%$) of these were unanswerable using the retrieved evidence.

\subsection{Analysis and Discussion}
\label{sec:exp-ablations}
We conducted analysis to assess baseline model performance under two conditions: (1) using ground-truth evidence (first block of Table~\ref{tab:exp-ablation-results}), and (2) disabling web-based evidence retrieval (second block of Table~\ref{tab:exp-ablation-results}). For both settings, we report conditional accuracy and justification generation scores with the evidence threshold set to $\lambda=0.3$.

\noindent\textbf{Baselines' Performance with Ground-truth Evidence. }
The results obtained using golden evidence represent upper-bound performance, highlighting the models’ full potential. 
Most baselines perform well in verdict prediction under this setting, underscoring both the importance and difficulty of effective evidence retrieval. 
Notably, the Gemini-based baseline achieves the highest scores for both prediction and justification generation. 
In contrast, baselines using LLaVA as the MLLM demonstrate the weakest performance. This is expected, as LLaVA~\citep{DBLP:journals/corr/abs-2407-07895} was not pre-trained on interleaved image-text documents. 
Across all baselines, we observe consistent failure in identifying conflicting claims. This result aligns with prior findings on the challenge of detecting conflicting textual claims~\citep{DBLP:conf/nips/SchlichtkrullG023}. Moreover, our dataset contains a relatively small number of conflicting claims, which can cause instability in model performance. 
It is also worth noting that our models do not explicitly model conflict within evidence, in contrast to~\citep{DBLP:conf/nips/SchlichtkrullG023}, which predicted verdicts per evidence piece and then examined whether these verdicts conflicted. 
In our setup, we observe stronger dependencies between multiple QA pairs that individual QA pairs are often insufficient to support or refute a claim. For example, only by combining the first and second QA pairs in Figure~\ref{fig:intro-exp} can the model conclusively refute the claim. 

Although baselines used ground-truth evidence, they still achieved low justification generation scores. We attribute this to the limitations of ROUGE-1 for evaluating justification generation. To address this, we experimented with a reference-based evaluation method, Ev2R~\citep{DBLP:journals/corr/abs-2411-05375}, which has been shown to align well with human assessments in open-ended generation tasks. Using this approach, the justification scores were much higher and more encouraging (details in Appendix~\ref{sec:app-add-exp-justi-scores}). Moving forward, we plan to explore and incorporate more appropriate evaluation methods for justification generation.

\noindent\textbf{Baselines' Performance without Searching. }
For the baseline without external evidence retrieval, we selected the combination of Qwen + Qwen-VL under PQG, as it demonstrated the strongest performance in question generation for claim verification. In this setting, all questions that require external search, those invoking the RIS, WST, or WSI tools, are marked with the response \textit{``No answer could be found.''}.
As shown in the second block of Table~\ref{tab:exp-ablation-results}, the model achieves an evidence score of $0.06$, reflecting a significant drop in evidence retrieval.
This result underscores the critical role of web-based information retrieval in real-world claim verification tasks. Nonetheless, in a few isolated cases, the model was still able to generate accurate evidence. These instances typically involved questions about locations or events depicted in the image, which the VQA tool could address correctly, likely because the model encountered these images during pretraining.
However, such behavior also raises concerns about potential data leakage from MLLMs.
Since the model's responses are not grounded in externally retrieved evidence, there is a risk that its predictions stem from memorized content or prior exposure to fact-checking articles during training.

\section{Limitations}
\label{sec:limitations}
AVerImaTeC has a relatively limited scale, as it is sourced from real-world claims and constructed through detailed human annotation. This is consistent with other human-annotated datasets of real-world claims, such as AVeriTeC~\citep{DBLP:conf/nips/SchlichtkrullG023}, which also contain only a few thousand claims.
Since the claims in AVerImaTeC originate from fact-checking articles, the dataset may inherit biases inherent in these sources—for example, selection bias~\citep{10.1111/jcom.12284,Barnoy10122019}, leading to imbalanced label distributions.

We made considerable efforts to prevent temporal leakage. Specifically, we provided annotators with custom search bars to retrieve online evidence published prior to the claim date. However, the exact dates of a small portion of claims ($5\%$) were unavailable. 
In such cases, we used the dates of the corresponding fact-checking articles, which may have been published a few days after the original claims. 
Additionally, we relied on Google Search and the Python package \textit{htmldate.find\_date} to estimate publication dates of web pages, which are coarse approximations. 

Furthermore, we exploited a reference-based evaluation strategy for both the generated questions and the retrieved evidence. Though it aligns well with human assessments, it has limitations in cases where model predictions are reasonable but not reflected in the reference annotations.  
In such cases, evaluation scores may be undeservedly low due to poor alignment with the references.

\section{Ethical Statement}
\label{sec:ethics}
The datasets and models described in this paper are not intended for use in truth-telling tasks, such as automated content moderation systems. The labels and justifications included in the dataset reflect only the evidence recovered by annotators and are therefore subject to the biases of both annotators and journalists. In addition, the QA annotations may introduce framing bias.

Annotators were instructed to prioritize evidence sources referenced in the original fact-checking articles, as we consider these sources, curated by professional fact-checkers, to be more trustworthy. Nonetheless, annotators were also permitted to draw on evidence retrieved by our customized search engine.
Using the list of common misinformative sources from~\citep{DBLP:conf/nips/SchlichtkrullG023}, we observed that 8 answers relied on a flagged source. However, this list was not applied during annotation, as it may be incomplete or contain false positives. Moreover, source credibility is often context-dependent, varying across topics and over time, which makes a static list insufficient for reliable filtering.

We acknowledge that some fact-checking articles included in our dataset may have been exposed to large pre-trained models during their pre-training phase. This is an open and ongoing challenge in constructing human-annotated datasets. To help mitigate this issue, the splits of our dataset are temporally organized. In particular, if the training data of a language model is cut off prior to the temporal start of our test set, then data leakage from pre-training into evaluation cannot occur.

We did not anonymize the data in AVerImaTeC, as all claims are derived from publicly available journalistic sources and primarily concern public figures and events. Preserving these references is essential to ensure the accuracy and integrity of fact-checking.
Nevertheless, we recognize that some individuals may not wish to appear in the dataset. Accordingly, we have established an opt-out policy: if any individual featured in the dataset, as a claim speaker, person depicted in an image, subject of a claim, or author of a fact-checking article underlying a claim, wishes to be removed, they may submit a request, and the relevant content will be deleted from the dataset.

\section{Conclusion}
\label{sec:conclusion}

We present a real-world image-text claim verification dataset, annotated with QA pairs that capture the reasoning and evidence retrieval processes involved in claim verification. 
To ensure high annotation quality, we employed a multi-stage evidence sufficiency validation process, resulting in substantial inter-annotator agreement on both verdicts and QA annotations.
In addition, we introduce a reference-based evaluation framework for open-web multimodal evidence retrieval, along with a set of baseline models for image-text claim verification that leverage web-sourced evidence.
These contributions provide a foundation for advancing research in image-text claim verification.

\clearpage

\section*{Acknowledgement}
This research was supported by the  Alan Turing Institute and DSO National Laboratories in Singapore Partnership (ref  DCfP2\textbackslash100063). Zifeng Ding and Andreas Vlachos were further supported by the
ERC grant AVeriTeC (GA 865958). Andreas Vlachos is also supported by the DARPA program SciFy.
Michael Schlichtkrull is supported by the Engineering and Physical Sciences Research Council (grant number EP/Y009800/1), through funding from Responsible AI UK (KP0016).

{
\small
\bibliography{ref}
\bibliographystyle{plainnat}
}

\section*{NeurIPS Paper Checklist}

\begin{enumerate}

\item {\bf Claims}
    \item[] Question: Do the main claims made in the abstract and introduction accurately reflect the paper's contributions and scope?
    \item[] Answer: \answerYes{} 
    \item[] Justification: Yes, main claims made in the abstract and introduction accurately reflect the 
paper’s contributions and scope.

\item {\bf Limitations}
    \item[] Question: Does the paper discuss the limitations of the work performed by the authors?
    \item[] Answer: \answerYes{} 
    \item[] Justification: Limitations of the work are discussed in Section~\ref{sec:limitations}.

\item {\bf Theory assumptions and proofs}
    \item[] Question: For each theoretical result, does the paper provide the full set of assumptions and a complete (and correct) proof?
    \item[] Answer: \answerNA{}
    \item[] Justification: There are no theoretical results in the paper.

    \item {\bf Experimental result reproducibility}
    \item[] Question: Does the paper fully disclose all the information needed to reproduce the main experimental results of the paper to the extent that it affects the main claims and/or conclusions of the paper (regardless of whether the code and data are provided or not)?
    \item[] Answer: \answerYes{} 
    \item[] Justification: The experiment settings and implementation details can be found in Section~\ref{sec:exp-baseline} (Model Implementation) and Appendix~\ref{sec:app-detail-model-imp}.

\item {\bf Open access to data and code}
    \item[] Question: Does the paper provide open access to the data and code, with sufficient instructions to faithfully reproduce the main experimental results, as described in supplemental material?
    \item[] Answer: \answerYes{} 
    \item[] Justification:  Both the data and code are accessible to reviewers (see Appendix~\ref{sec:data_stat}).  As we anticipate using the dataset in a future shared task, we are as of submission
time only releasing the training and development splits. We will make the test split available privately
to reviewers upon request. We will make our data and code publicly available and maintain them on GitHub upon acceptance. 

\item {\bf Experimental setting/details}
    \item[] Question: Does the paper specify all the training and test details (e.g., data splits, hyperparameters, how they were chosen, type of optimizer, etc.) necessary to understand the results?
    \item[] Answer: \answerYes{} 
    \item[] Justification:
    \item[] Guidelines: The information for data splits can be found in Section~\ref{sec:data_stat}. Testing details are provided in Appendix~\ref{sec:app-detail-model-imp}.
    \begin{itemize}
        \item The answer NA means that the paper does not include experiments.
        \item The experimental setting should be presented in the core of the paper to a level of detail that is necessary to appreciate the results and make sense of them.
        \item The full details can be provided either with the code, in appendix, or as supplemental material.
    \end{itemize}

\item {\bf Experiment statistical significance}
    \item[] Question: Does the paper report error bars suitably and correctly defined or other appropriate information about the statistical significance of the experiments?
    \item[] Answer: \answerNA{} 
    \item[] Justification: Our goal is not to advance in model design but to establish baselines for image-text claim verification. Besides, following the practice of~\citep{DBLP:conf/nips/SchlichtkrullG023} and prior FEVER workshops (https://fever.ai/workshop.html), we do not require multiple runs of each model.

\item {\bf Experiments compute resources}
    \item[] Question: For each experiment, does the paper provide sufficient information on the computer resources (type of compute workers, memory, time of execution) needed to reproduce the experiments?
    \item[] Answer: \answerYes{} 
    \item[] Justification: The information for computation resources can be found in Appendix~\ref{sec:app-detail-model-imp}.
    
\item {\bf Code of ethics}
    \item[] Question: Does the research conducted in the paper conform, in every respect, with the NeurIPS Code of Ethics \url{https://neurips.cc/public/EthicsGuidelines}?
    \item[] Answer: \answerYes{} 
    \item[] Justification: The research in the paper conforms, in every respect, with the NeurIPS Code of Ethics.

\item {\bf Broader impacts}
    \item[] Question: Does the paper discuss both potential positive societal impacts and negative societal impacts of the work performed?
    \item[] Answer: \answerYes{} 
    \item[] Justification: The potential societal impact of the work has been discussed in Section~\ref{sec:intro} and Section~\ref{sec:related}.
    
\item {\bf Safeguards}
    \item[] Question: Does the paper describe safeguards that have been put in place for responsible release of data or models that have a high risk for misuse (e.g., pretrained language models, image generators, or scraped datasets)?
    \item[] Answer: \answerNA{} 
    \item[] Justification: The paper poses no such risks.

\item {\bf Licenses for existing assets}
    \item[] Question: Are the creators or original owners of assets (e.g., code, data, models), used in the paper, properly credited and are the license and terms of use explicitly mentioned and properly respected?
    \item[] Answer: \answerYes{} 
    \item[] Justification: We used partially annotated data from AVeriTeC, AMMeBA and ClaimReview. They are explictly mentioned and discussed in Appendix~\ref{sec:app-state-dataset}

\item {\bf New assets}
    \item[] Question: Are new assets introduced in the paper well documented and is the documentation provided alongside the assets?
    \item[] Answer: \answerYes{} 
    \item[] Justification: 
     The dataset and code have been made available to reviewers.
    Our dataset and baseline will be made publicly 
available under a CC-BY-NC-4.0 license if accepted. Detailed dataset information can be found in Section~\ref{sec:anno_struct},~\ref{sec:anno_pipe} and~\ref{sec:data_stat}.

\item {\bf Crowdsourcing and research with human subjects}
    \item[] Question: For crowdsourcing experiments and research with human subjects, does the paper include the full text of instructions given to participants and screenshots, if applicable, as well as details about compensation (if any)? 
    \item[] Answer: \answerYes{} 
    \item[] Justification: Data collection was carried out with the help of Appen (https://appen.com/). The payment details and annotator demographics can be found in Appendix~\ref{sec:app-anno-detail-annotator-demo}. Guidelines for annotators are provided Appendix~\ref{sec:app-annotation-guidelines}.

\item {\bf Institutional review board (IRB) approvals or equivalent for research with human subjects}
    \item[] Question: Does the paper describe potential risks incurred by study participants, whether such risks were disclosed to the subjects, and whether Institutional Review Board (IRB) approvals (or an equivalent approval/review based on the requirements of your country or institution) were obtained?
    \item[] Answer: \answerNA{} 
    \item[] Justification: We do not involve human subjects. All annotators are recruited by a company, Appen (See Appendix~\ref{sec:app-anno-detail-annotator-demo}).

\item {\bf Declaration of LLM usage}
    \item[] Question: Does the paper describe the usage of LLMs if it is an important, original, or non-standard component of the core methods in this research? Note that if the LLM is used only for writing, editing, or formatting purposes and does not impact the core methodology, scientific rigorousness, or originality of the research, declaration is not required.
    \item[] Answer: \answerNA{} 
    \item[] Justification: The core method development in this research does not involve LLMs as any important, original, or non-standard components. LLMs are only used for grammar error correction.

\end{enumerate}

\clearpage
\appendix
\section{Statement of Dataset}
\label{sec:app-state-dataset}
We provide the training and development splits of our dataset for review.\footnote{https://kaggle.com/datasets/a1ebcc27f233ca16e80fa7007d1a8cea6c491656d7b4ca6d87531df61c5089fd}
However, due to naming inconsistencies with Kaggle, we are unable to upload the images there. Instead, we have made the image files available on Harvard Dataverse.\footnote{https://dataverse.harvard.edu/previewurl.xhtml?token=5fbd9e3c-bc9e-49d7-a109-eace6f1bdacd}  
We cannot upload the entire dataset (including the .json file and images) to Harvard Dataverse, as it requires publishing the dataset when generating the Croissant file.
To avoid unexpected inaccessibility of images on Harvard Dataverse, we also upload the whole development set (images and json file) to \textbf{supplementary}, due to file size limits.

Since we plan to use the test split for future shared tasks, we are withholding it at the time of submission. However, the test split will be made privately available to reviewers upon request.

If accepted, we will publicly release the training and development splits of the dataset and maintain both the data and baseline code on Harvard Dataverse and GitHub. The dataset and baselines will be licensed under the CC BY-NC 4.0 license.

\section{Annotation Platform and Guidelines}
\label{sec:app-annotation-guidelines}
The annotation was performed using a custom-developed web platform specifically designed for this task from our team. We will make the platform’s source code available upon request.

Annotators were provided with detailed guidelines outlining the annotation process.
Due to the length of the guideline, we attached it to the \textbf{supplementary}.
After reviewing the instructions, they underwent training for each phase. 
Specifically, we provide $14$, $12$ and $12$ training instances for the first three phases, respectively. Phases four and five replicate phases two and three; therefore, no additional training instances were provided.
We offered continuous feedback to annotators throughout both the training and annotation phases.

\section{Annotation Fees and Annotator Demographics}
\label{sec:app-anno-detail-annotator-demo}
The annotation was conducted with the assistance of Appen,\footnote{https://www.appen.com/} a private company that provides machine learning services. Annotators were recruited through the company, which ensures fair compensation practices in accordance with their fair pay guidelines.\footnote{https://success.appen.com/hc/en-us/articles/9557008940941-Guide-to-Fair-Pay}  
A total of 20 annotators participated in the project, 12 women ($60\%$) and 8 men ($40\%$).
Fourteen annotators were based in the United States and six in the United Kingdom.
Regarding age distribution,  $20\%$ of annotators were between 18 and 30 years old, $35\%$ between 31 and 45, $25\%$ were older than 45, and $20\%$ chose not to disclose their age.

\section{Additional Dataset Information}
\label{sec:app-data-info}
In this section, we provide additional dataset statistics. 
\subsection{Statistics for Data Filtering in Phase One}
\label{sec:app-data-p1-data-filter}
We began with $2,353$ fact-checking articles in Phase One (P1) of the annotation process. To increase the proportion of image-text claims, we incorporated partially annotated articles from three sources:
1) filtered articles from AVeriTec~\citep{DBLP:conf/nips/SchlichtkrullG023} containing multimodal claims,  2) articles verifying image-related claims from AMMEBA~\citep{DBLP:journals/corr/abs-2405-11697} and 3) all \textit{true} claims from ClaimReview\footnote{https://developers.google.com/fact-check/tools/api} over the past two years that included keywords such as \textit{phtos} or \textit{pictures}.
Among the 2,353 articles, we identified 91 duplicate claims (e.g., those verified by multiple fact-checking organizations).
Duplicates were detected automatically by computing cosine similarity and Jaccard distance between N-gram representations of textual part of claims.
We manually reviewed claim pairs with a cosine similarity or Jaccard distance greater than 0.3. 
Despite these efforts, a small number of duplicates (fewer than 5\%) remained, as fact-checkers may paraphrase similar claims differently. 
However, due to the temporal split of our dataset, such duplication does not cause data leakage across the splits. Paywalled or inaccessible articles were reported by annotators and subsequently discarded.

Although we leveraged coarsely annotated articles that verify multimodal claims, not all of them involved image-text relationships.
In the first step, we filtered out claims of other modalities and retained only the image-text claims (around $85\%$) for further P1 annotation, which included image uploading, metadata collection, and verdict annotation.
We further filter out image-text claims, where images are not used in claim verification (14.7\%) or which involve image manipulations(4.0\%). Both types were excluded from subsequent phases.
Following~\citep{DBLP:conf/nips/SchlichtkrullG023}, we discard \textit{speculative} claims, personal opinions and claims relying solely on \textit{fact-checking reference} as the fact-checking strategy.
After the P1 annotation, we retained a total of $\mathbf{1,457}$ valid image-text claims, for further processing in subsequent phases.

\subsection{Discarded Claims}
\label{sec:app-discard-claims}
In Phase Two, annotators can report and skip invalid claims, particularly when images associated with claims are inaccessible due to regional restrictions. A total of $48$ such cases were discarded.

As described in Section~\ref{sec:anno_pipe}, 
when there are conflicting verdicts between annotators in Phases Two and Three, the affected claims proceed to Phases Four and Five for quality assurance updates and a second round of evidence sufficiency checks. In our dataset, $14.6\%$ of claims required this second round of evaluation.

Claims with unresolved conflicts after the five-phase annotation process are also discarded. Overall, approximately $7\%$ of claims were excluded due to irreconcilable disagreements in annotated verdicts, even after the second round of evidence sufficiency checks.

\subsection{Metadata Distributions}
\label{sec:app-data-statistics-metadatao}
Following~\citep{DBLP:conf/nips/SchlichtkrullG023}, we provide details of distributions of metadata of data splits in this part.

\begin{table*}[ht!]
\centering
\caption{\textbf{Distributions of claim types.} }
  \label{tab:app-meta-claim-type}
  \begin{tabular}{lccc}
    \toprule
    \textbf{Claim Type} &\textbf{Train} &\textbf{Dev} &\textbf{Test} \\
    \midrule
    Event/Property Claim &85.4 &91.4 &93.5\\
    Causal Claim  & 5.7&0.7 & 3.4\\
    Numerical Claim   & 3.2&3.9 &1.4 \\
    Media Analysis Claim  &21.1&14.5 &30.1 \\
    Media Publishing Claim    & 2.1& 2.0&2.0 \\
    Position Statement  & 1.2&0.7 & 0.3\\
    Quote Verification  & 1.4& 0.7& 0.3\\
    \bottomrule
\end{tabular}
\end{table*}

\begin{table*}[ht!]
\centering
\caption{\textbf{Distributions of fact-checking strategies.} }
  \label{tab:app-meta-factcheck}
  \begin{tabular}{lccc}
    \toprule
    \textbf{Claim Type} &\textbf{Train} &\textbf{Dev} &\textbf{Test} \\
    \midrule
    Media Source Discovery &26.4 &20.4 &26.4 \\
    Keyword Search &19.5 & 19.7 &14.8 \\
    Written Evidence  & 87.5& 89.5&84.7 \\
    Reverse Image Search  &50.2 & 57.9 & 67.3\\
    Image Analysis  &20.7 & 24.3 & 21.3\\
    Consultation   &30.0 &24.3 & 16.8\\
    Video Analysis   & 6.8&4.6 & 8.2\\
    Numerical Comparison   & 2.9& 3.3&1.4 \\
    Satirical Source Identification  &1.9 & 3.3&1.4 \\
    Fact-checker Reference   & 11.0&11.8 &3.7 \\
    Geolocation  & 4.0&5.3&5.4\\
    \bottomrule
\end{tabular}
\end{table*}

\noindent\textbf{Metadata about Claims. }We  present the distribution of claim types and fact-checking strategies in Table~\ref{tab:app-meta-claim-type} and~\ref{tab:app-meta-factcheck}, respectively. It is important to note that a single claim may belong to multiple claim types and can be verified using several fact-checking strategies.
\begin{table*}[ht!]
\centering
\caption{\textbf{Counts of locations associated with claims.} Countries are represented with their ISO country code. We do not show countries with fewer than ten occurrences in the table, whereas the complete location information is available in our dataset.}
  \label{tab:app-meta-location}
  \begin{tabular}{lccc}
    \toprule
    \textbf{Country code} &\textbf{Counts} \\
    \midrule
    IN & 417 \\
    US & 309 \\
    GB & 74 \\
    UA &37 \\
    PK & 30 \\
    IL &27\\
    NG &20\\
    LK &19 \\
    TR &18 \\
    PS &17 \\
    KR &17\\
    KE &15\\
    AU &14 \\
    RU &13 \\
    BD &12 \\
    JP &11 \\
    TH &10 \\
    MY & 10 \\
    CN &10 \\
    CA &10\\
    CR &10 \\
    \bottomrule
\end{tabular}
\end{table*}
The most relevant locations associated with claims are listed in Table~\ref{tab:app-meta-location}.
We observed a bias in the geographic distribution of claims. As our dataset includes claims verified by fact-checkers, it may inherit any biases present in the original fact-checking sources~\citep{10.1111/jcom.12284,Barnoy10122019}. 

\begin{table*}[ht!]
\centering
\caption{\textbf{Distributions of question types.} }
  \label{tab:app-meta-qt}
  \begin{tabular}{lccc}
    \toprule
    \textbf{Question Type} &\textbf{Train} &\textbf{Dev} &\textbf{Test} \\
    \midrule
    Text-related & 35.2 & 38.9& 31.8 \\
    Image-related &63.4  & 58.8&62.2  \\
    Metadata-related  & 3.2 &4.4 & 8.3 \\
    Commonsense-related & 0.8 &0.9 & 1.1   \\
    \bottomrule
\end{tabular}
\end{table*}

\begin{table*}[ht!]
\centering
\caption{\textbf{Distributions of metadata related to answers.} The first block is the distribution for answering methods, the second for answer types and the third for source medium.}
  \label{tab:app-meta-ans}
  \begin{tabular}{lccc}
    \toprule
    \textbf{Question Type} &\textbf{Train} &\textbf{Dev} &\textbf{Test} \\
    \midrule
    Image-search &54.6  & 54.6&51.9\\
    Text-search &40.1&41.0 &37.4\\
    Metadata  & 1.8 &3.0 &7.7\\
    Image Analysis &3.2  & 1.4&2.5\\
    \midrule
    Abstractive & 16.4 &15.0 & 19.1 \\
    Extractive &  57.9&61.3 & 58.9\\
    Unanswerable &2.2  &3.2 & 3.2 \\
    Boolean &  21.9& 18.8& 17.3 \\
    Image & 1.6 &1.6 &1.5\\
    \midrule
    Web text &  85.2&85.2 &78.9\\
    PDF & 0.8 &1.2 & 0.8 \\
    Metadata & 1.2 & 2.1& 7.1 \\
    Video &  2.5&1.2 &2.9  \\
    Image/graphic & 4.4 & 5.3& 4.6 \\
    Web table & 0.3 & 0.5& 0.1 \\
    Other & 0.2 &0.0 & 0.1 \\
    \bottomrule
\end{tabular}
\end{table*}
\noindent\textbf{Metadata about QA Annotations. }
In addition to metadata for claims, we also provide metadata associated with the annotated QA pairs, including question types, answering methods, answer types and source mediums.
Metadata statistics for questions and answers are presented in Table~\ref{tab:app-meta-qt} and Table~\ref{tab:app-meta-ans}, respectively.
Notably, a single question may belong to multiple question types.

\section{Inter-Annotator Agreement}
\label{sec:app-inter-annotator-agreement}
\begin{figure*}[ht!] 
	\centering
	\includegraphics[width=\linewidth]{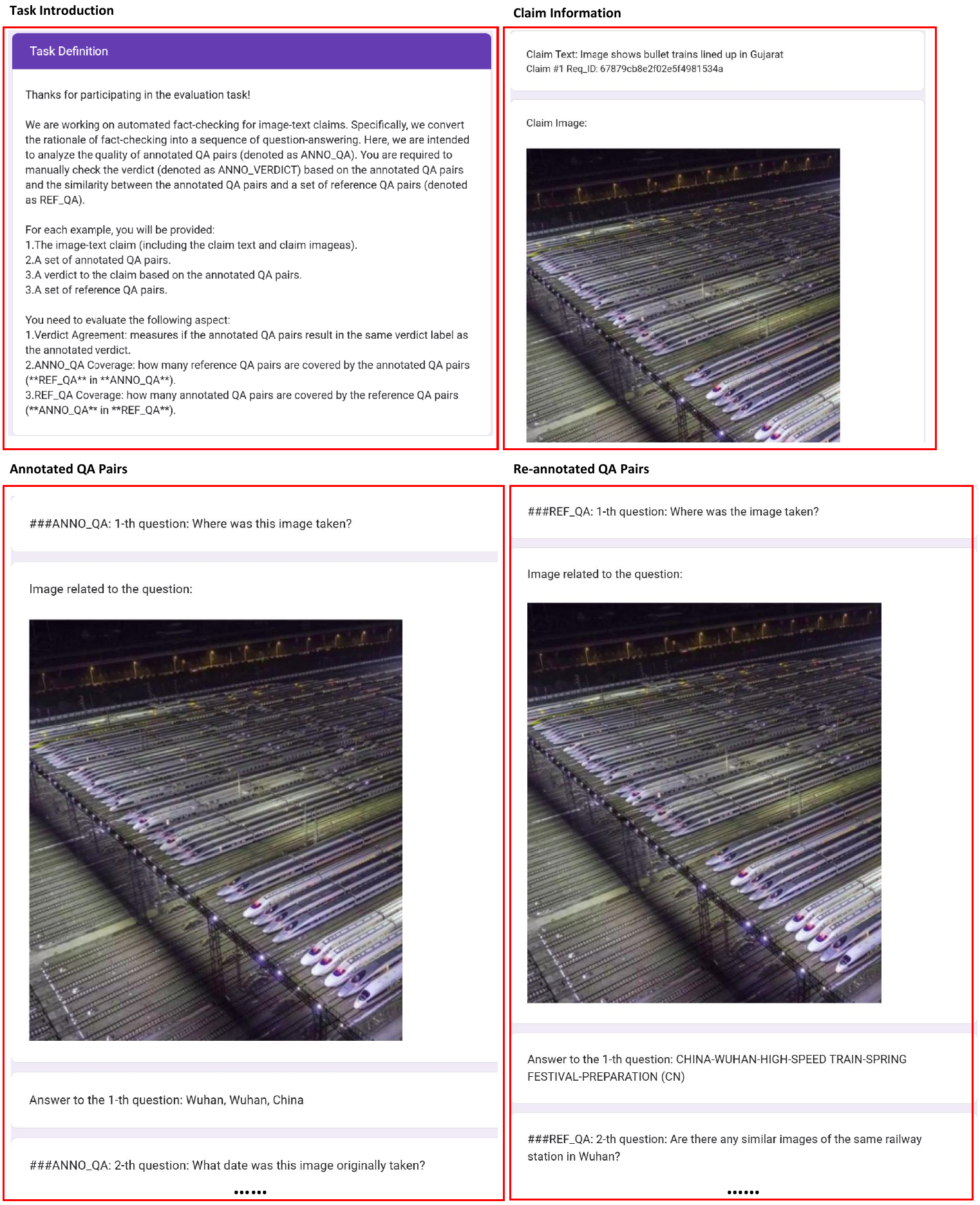} 
	\caption{
 \textbf{Platform and instructions for validating annotators' agreement on QA annotations.}}
	\label{fig:app-human-eval-inter}
\end{figure*}
To assess the quality of our annotated data, we recruited a different set of annotators to re-annotate $100$ randomly sampled claims from our dataset and performed an inter-annotator agreement check. 
During re-annotation, we assumed that claim extraction and normalization had already been completed, and the annotators proceeded with the remaining phases. We ensured that the sample included at least five claims for each veracity label.
The inter-annotator agreement check was done regarding both the agreement on verdicts and agreements on annotated QA pairs. 

\noindent\textbf{Verdict Agreement. }
To evaluate agreement on verdicts, we used Randolph's~\citep{10.1007/s11634-010-0073-4} free-marginal multi-rater $\kappa$,  which is well-suited for unbalanced datasets, following previous practices~\citep{DBLP:conf/nips/SchlichtkrullG023,DBLP:conf/emnlp/OusidhoumY022}. We achieved an agreement score of 0.742 on the double-annotated claims.

\noindent\textbf{QA Pair Agreement. }
For assessing agreement on annotated QA pairs, we recruited three best performing annotators to compare the similarities between the original and re-annotated QA annotations. Specifically, we instructed them to evaluate  1) whether the annotated verdict from our dataset could be supported by the original QA pairs; 2) how many original QA pairs for a claim are covered by the re-annotated QA pairs, and 3) how many QA pairs from the re-annotation are covered by the original QA pairs. The platform used for this agreement check is shown in Figure~\ref{fig:app-human-eval-inter}.

\noindent\textbf{Justification Evaluation. }
Although we primarily focused on verdict and QA annotation agreement, we also conducted a small-scale human evaluation to assess the quality of justifications. We randomly sampled 20 justifications from our dataset and from~\citep{DBLP:conf/nips/SchlichtkrullG023}, and asked a human evaluator to rate them on a scale from 0 to 5. Our justifications achieved an average score of 4.2, compared to an average score of 1.95 from~\citep{DBLP:conf/nips/SchlichtkrullG023}.

\section{Human Alignment of Evaluation Metrics}
\label{sec:app-human-align-eval-metrics}
Following~\citep{DBLP:conf/nips/SchlichtkrullG023}, we evaluate the quality of generated questions and retrieved evidence. Motivated by a recent study~\citep{DBLP:journals/corr/abs-2411-05375},
we adopt a reference-based evaluation method that compares model responses to human-annotated ground-truth data. This method is applied to both question and evidence evaluations.
To assess the reliability of the reference-based evaluation, we compare the resulting scores with human judgments obtained from independent raters (see Appendix~\ref{sec:app-eval-ques-eval} and \ref{sec:app-eval-evid-eval}). 
Additionally, we conduct \textit{checklist} tests to evaluate the sensitivity of the reference-based evidence evaluation method, following the approach outlined in~\citep{DBLP:journals/corr/abs-2411-05375} (Appendix~\ref{sec:app-eval-robust}).

\subsection{Alignment Check on Question Evaluation. }
\label{sec:app-eval-ques-eval}
Regarding question evaluation, we focus on assessing the semantics (i.e., textual content) of the questions. This evaluation approach aligns with that used in~\citep{DBLP:journals/corr/abs-2411-05375} and has also been applied at the FEVER workshop,\footnote{https://huggingface.co/spaces/fever/AVeriTeCFEVER8} demonstrating its reliability and strong correlation with human ratings.

To further ensure alignment with human judgment in our setting, we conducted a small-scale human evaluation to validate the reference-based evaluation of questions. Specifically, we invited two NLP researchers (both authors of this paper) to participate in the assessment. They were presented with model-generated questions from the first 20 testing claims in our dataset and were asked to compare these questions against ground-truth annotated questions. The comparison was made based on two criteria:
\textbf{Relevance} to claim verification (how relevant the generated questions are for verifying the claim) and \textbf{Coverage} of the ground-truth questions (how many ground-truth questions are accurately covered by the predicted questions.).
The agreement between human and the evaluation metric was measured using 
Spearman ($\rho$)~\citep{spearman04}, achieving a value of \textbf{0.705}, and Pearson correlation coefficients ($r$)~\citep{pearson1896mathematical} achieving a value of  \textbf{0.791}. These results indicate a strong alignment between human judgments and the reference-based evaluation scores on question evaluation.

\begin{table*}[ht!]
\centering
\caption{\textbf{Correlation between human evaluation and automatic evaluation metrics.} We presents both the Spearman ($\rho$) and Pearson ($r$) correlation coefficients.}
  \label{tab:app-align-evid-eval}
  \begin{tabular}{lccc}
    \toprule
    \textbf{Scorer} &\textbf{Text-only} &\textbf{Interleaved} &\textbf{Separate} \\
    \midrule
    $\rho$ &0.263 &0.08 &0.332 \\
    $r$ &0.286 & 0.14&0.381\\
    \bottomrule
\end{tabular}
\end{table*}

\subsection{Alignment Check on Evidence Evaluation. }
\label{sec:app-eval-evid-eval}
We leveraged re-annotation and re-used the human judgment on QA annotations as described in Appendix~\ref{sec:app-inter-annotator-agreement} to validate the alignment between our automatic evidence evaluation method and human judgment. 
To assess alignment, we treated the original annotations from our dataset as predictions and the re-annotations as ground truth. We then applied the automatic evaluation methods to compare these \textit{``predictions''} against the ground truth. Subsequently, we computed alignment scores by comparing the recall obtained from the automatic evaluation methods with human annotations.
The correlation between the automatic evaluation scores and human assessments is presented  using Spearman ($\rho$) and Pearson ($r$) correlation coefficients, as shown in Table~\ref{tab:app-align-evid-eval}.
Our analysis shows that the separated evaluation method aligns most closely with human assessments, demonstrating the effectiveness of our approach.

\begin{table*}[ht!]
\centering

\caption{\textbf{Adversarial attack results for the assessment of the robustness of our evaluation method.} Results in the table are obtained by computing the evidence evaluation score difference (in \%) between initial evidence and
manipulated evidence with checklists.}
  \label{tab:app-checklist-evid-eval}
  \begin{tabular}{lccc}
    \toprule
    \textbf{Check.} &\textbf{Text-only}&\textbf{Interleaved} &\textbf{Separate} \\
    \midrule
    Completeness & -29.2&-19.8& -28.5 \\
    Shuffle &-10.2 &4.1& -17.9 \\
    \textbf{Irrelevant Image} & 0.0&9.4 &-32.7\\
    \midrule
    Inv. contract &-1.6 & -7.4&1.5 \\
    Inv. num2text  & 1.8&21.3 &-11.3 \\
    Inv. text2num  & -0.3&24.5 &0.6 \\
    Inv. synonyms  &-11.8 &88.8 &-0.6\\
    Redundant words  & -3.1& 85.4&-4.2 \\
    Fluency  &-4.9& 31.4& -0.7\\
    Argument structure  & -1.6& 30.8& 4.8\\
    \textbf{Image Invariance} & 0.0&20.0 & -1.4 \\
    \bottomrule
\end{tabular}
\end{table*}

\subsection{Robustness Check on Evidence Evaluation. }
\label{sec:app-eval-robust}
Motivate by~\citep{DBLP:journals/corr/abs-2411-05375,ribeiro-etal-2020-beyond}, we not only assess alignment but also validate the robustness of our evidence evaluation method using \textit{checklist} tests. Notably, the sensitivity of reference-based evaluation for textual data has already been examined in~\citep{DBLP:journals/corr/abs-2411-05375}, and we adopt their adversarial attack design on texts. Details for these attacks (\textit{completeness}, \textit{shuffle}, \textit{Inv. contract}, \textit{Inv. num2text}, \textit{Inv. text2num}, \textit{Inv. synonyms}, \textit{Redundant words}, \textit{Fluency}, \textit{Argument structure}) can be found in the original study.

Given the multimodal nature of our evaluation, we extend the robustness tests to include visual adversarial checks. Specifically, we evaluate the robustness of the evidence evaluation method against two types of image-based perturbations: 1) \textbf{Irrelevant images}, replacing  images in predicted evidence with unrelated images and 2) \textbf{Image invariance}, applying invariant manipulations, such as resizing and rotation, to images within the predicted evidence

We conducted robustness testing on $20$ claims and compared the difference between the original evidence evaluation scores and the scores obtained after introducing adversarial evidence. The results are summarized in
Table~\ref{tab:app-checklist-evid-eval}. 
A robust evaluation method is expected to show a significant performance drop when facing adversarial attacks in the first block of the table while maintaining consistent scores (i.e., minimal deviation from previous scores) when subjected to attacks in the second block.

The results indicate that both the Text-only and Separated reference-based evaluations exhibit robustness against textual adversarial attacks. In contrast, the interleaved evaluation is sensitive and unstable when faced with such attacks. This finding suggests that even advanced MLLMs, such as Gemini, may be prone to instability when handling interleaved image-text comprehension.

Regarding visual adversarial attacks (i.e., irrelevant image replacement and image invariance), the Text-only evaluation method fails to maintain stability, while our Separated evaluation method demonstrates robust performance.

\section{Additional Experimental Results}
\label{sec:app-add-exp-results}

\subsection{Zero-shot Performance of Baselines}
\label{sec:app-add-exp-zero-shot}
\begin{table*}[ht!]
\centering
\caption{\textbf{Zero-shot performance of baselines.} \textbf{Q-Eval} and \textbf{Evid-Eval} denote for \textit{recall} scores of generated questions and retrieved evidence, with reference of ground-truth questions and evidence. We report verdict prediction and justification generation scores conditioned on evidence retrieval performance, specifically only considering verdict accuracy and justification generation performance when the evidence score is above 0.2, 0.3 and 0.4.}
  \label{tab:app-add-exp-result-main}
  \begin{tabular}{ll|cc|ccc|ccc}
    \toprule
    \textbf{LLM} &\textbf{MLLM} &\textbf{Q-Eval} &\textbf{Evid-Eval} & \multicolumn{3}{c}{\textbf{Veracity} \@ (.2/.3/.4)} & \multicolumn{3}{c}{\textbf{Justifications} \@ (.2/.3/.4)}\\
    \midrule
    \rowcolor{lightgray}
   \multicolumn{10}{c}{\textit{Paralleled Question Generation}}\\ 
    \midrule
    Gemini & Gemini &0.40 &0.18 &0.13&0.11 &0.08 &0.14 &0.13 & 0.08\\
    Qwen & Qwen-VL &0.41 &0.16 &0.08 &0.07 & 0.04&0.11&0.09 & 0.07\\
    Gemma & Gemma  &0.37 &0.22 &0.12 & 0.12& 0.08& 0.17&0.15 & 0.10\\
    Qwen & LLaVA  & 0.37&0.18 &0.10 &0.09 & 0.07&0.12 &0.11 &0.07 \\
    \midrule
    \rowcolor{lightgray}
   \multicolumn{10}{c}{\textit{Dynamic Question Generation}}\\ 
    \midrule
     Gemini & Gemini &0.35 &0.19 & 0.15& 0.14&0.09 &0.15 &0.13 & 0.10\\
    Qwen & Qwen-VL &0.30 &0.14 & 0.11&0.10&0.06 &0.11 &0.09 & 0.06\\
    Gemma & Gemma  & 0.27& 0.19&0.12 & 0.11&0.07 &0.14& 0.13& 0.08\\
    Qwen & LLaVA  & 0.24&0.16 & 0.11&0.11 &0.06 & 0.12&0.11 & 0.06\\
    \midrule
    \rowcolor{lightgray}
   \multicolumn{10}{c}{\textit{Hybrid Question Generation}}\\ 
    \midrule
     Gemini & Gemini &0.35 & 0.16&0.13 &0.12 & 0.09&0.12 &0.11 & 0.08\\
    Qwen & Qwen-VL &0.34 &0.14 &0.09 &0.08 &0.05 &0.10 & 0.09&0.05 \\
    Gemma & Gemma  & 0.26& 0.25& 0.15& 0.14& 0.09& 0.19&0.18 &0.12 \\
    Qwen & LLaVA  &0.26 &0.17&0.07 & 0.07& 0.05&0.12 &0.11 &0.08 \\
    \bottomrule
\end{tabular}
\end{table*}

Due to the limitation of space, we provide the zero-shot performance of baselines in Table~\ref{tab:app-add-exp-result-main}. The findings discussed in Section~\ref{sec:exp-results} hold for baselines under the zero-shot setting as well.

\begin{table*}[ht!]
\centering
\caption{\textbf{Justification evaluation scores with ROGUE-1 and Ev2R } when baselines are using ground-truth evidence.}
  \label{tab:ev2r-justi}
  \begin{tabular}{ll|c|c}
    \toprule
    \textbf{LLM} &\textbf{MLLM} &\textbf{ROGUE-1} &\textbf{Ev2R} \\
    \midrule
    Gemini & Gemini &0.50 &0.78 \\
    Qwen & Qwen-VL  &0.44 &0.68 \\
    Gemma & Gemma  &0.43 &0.75 \\
    Qwen & LLaVA  & 0.49& 0.55\\
    \bottomrule
\end{tabular}
\end{table*}
\subsection{Justification Generation Scores with Ev2R}
\label{sec:app-add-exp-justi-scores}
The justification evaluation scores of the baselines, even when provided with ground-truth evidence, are relatively low, as shown in Table~\ref{tab:exp-ablation-results}. We attribute this to the limitations of ROUGE-1 in assessing open-ended generation. To address this, we adopted Ev2R~\citep{DBLP:journals/corr/abs-2411-05375}, a reference-based evaluation method shown to perform well for open-ended generation tasks. The corresponding results are reported in Table~\ref{tab:ev2r-justi}.

A comparison between ROUGE-1 and Ev2R reveals that Ev2R produces significantly higher and more reasonable scores, which we find encouraging. As part of future work, we plan to conduct human alignment studies and incorporate such more sophisticated evaluation approaches for justification generation.


\section{Details of Experiments}
\label{sec:app-detail-of-exp}

\subsection{Model Implementation}
\label{sec:app-detail-model-imp}
\noindent\textbf{Hyper-parameters and Implementation. }
In the main experiments, we set the total number of generated questions to be $5$ for all QG strategies. For HQG, the first two questions are generated in parallel while the rest three are generated exploited DQG.
In cases of textual evidence retrieval (i.e., leveraging the tools of RIS and WST), we truncated retrieved texts into chunks with the maximum length of $128$ and applied BM25~\citep{DBLP:journals/ftir/RobertsonZ09} to select the most relevant chunks to the given query. 
With the increasing capability of long-context understanding of existing LLMs, we keep the top $30$ most related chunks, without a second stage of fine-grained re-ranking as what previous works have done~\citep{DBLP:conf/naacl/ChenKSDC24,DBLP:conf/nips/SchlichtkrullG023}.
For the retrieved images returned by WSI, we compute their similarity scores with the given textual query with CLIP~\citep{DBLP:conf/icml/RadfordKHRGASAM21} and select the most related one as the image evidence source.

For the choice of LLMs in baselines, we have tried using LLaMA-3.1-8B-Instruct~\citep{llama-meta} as the LLM, whereas the model got stuck in loops, the same as reported by other users.\footnote{https://www.reddit.com/r/LocalLLaMA/comments/1c858ac/llama3\_seems\_to\_get\_stuck\_in\_loops\_sometimes/} 
For LaVA-Next~\citep{DBLP:journals/corr/abs-2407-07895}, which is not designed for interleaved image-text, we only consider the textual part of evidence in verdict prediction and justification as we observed some issues with model generation with complex interleaved image-text information.

For the searching related tools, specifically WST, WSI and RIS, we used the API provided by Google. For web search with textual queries (WST and WSI), we first tokenize and post-tag words in queries and only keep verbs, nouns and adjectives as the search term~\citep{DBLP:conf/ranlp/KaradzhovNMBK17}. 
We set temporal constraints with input arguments, limiting all returned web pages published before claim dates. We keep the first 30 search results. 
For RIS, we employed the google cloud vision service for detecting web pages containing matched images with the querying image. However, the service does not embed arguments to set temporal constraints. Alternatively, we use a post-hoc method by leveraging the Python package \textit{htmldate.fine\_date} to filter out pages published before claim dates. We noticed a lot of web pages returned by RIS are social media posts, which are non-scrapable. For these pages, we use their page titles as the scraping content.

\noindent\textbf{Few-shot Setting. }To encourage models generate more critical questions for fact-checking, we exploit a computationally efficient method, few-shot learning. Specifically, we use a few training examples to guide models in question generation. Selecting similar claims to the inference one is important as similar claims may have similar reasoning path for claim verification. We rank the similarity between training claims and the inference claim with BM25~\citep{DBLP:journals/ftir/RobertsonZ09} by comparing their textual part. We set the number of shots to be $3$ to balance between the input length and information from demonstrations. 

For the PQG strategy, we directly provide models the ground-truth questions from selected demonstrations. For the DQG setting, models are provided the textual part of image-text claims and their first questions to generate the initial questions. For generating subsequent questions, each demonstration contains the textual part of a claim, its evidence history from previous QA pairs and the next question to be asked.

\noindent\textbf{Guidance for Tool Selection. }Besides the guidance for tool selection in prompts as provided in Appendix~\ref{sec:app-prompt-tool-sel}, we also used few demonstrations for inspire models to select proper tools.
We provide few examples to guide tool selection as in the preliminary experiment we observed a heavy rely on VQA as the answering tool. We leveraged the metadata annotation of QA pairs, the answering method, for tool selection. \textit{Image-search} will be mapped to RIS, \textit{Text-search} with an image answer will be converted to WSI while with a textual answer will be mapped into WST. \textit{Image analysis} will be converted to selecting VQA as the answering tool.

\noindent\textbf{Computation Resources. }
All experiments are conducted with two GPUs each with 40G dedicated memory.
Specifically, we exploited either A100 or L40 for our experiments.
The Qwen + Qwen-VL baselines and the Qwen + LLaVA baselines take about three hours with A100 and Gemma-based baselines take about seven hours for inference on the test split.
Models have a faster inference speed on L40, saving one third of inference time.
The inference time of Gemini-based models varies, probably depending on the volume of API calls. 
Also, we observed instability of Gemini API (e.g., the API call returned \textit{503}, saying the service is not available), maybe because of too many requests at the same time.

\subsection{Experiment Environment and Packages}
\label{sec:app-detail-exp-setting}
In this section, we introduce the experiment environment and packages in use.
We implement all models under the
PyTorch Library (version 2.4.0+cu121), with CUDA version 12.1. For the implementation of open-source LLMs and MLLMs, we leverage the HuggingFace Library, with the \textit{Qwen/Qwen2.5-7B-Instruct} (Qwen), \textit{google/gemma-3-12b-it} (Gemma), \textit{Qwen/Qwen2.5-VL-7B-Instruct} (Qwen-VL) and \textit{llava-hf/llava-v1.6-mistral-7b-hf} (LLaVA), respectively. The version of Huggingface is 4.50.2. For Gemini, we exploited the API, \textit{gemini-2.0-flash-001}. For the CLIP model employed in image-text similarity computation for evidence rankin, we adopt the checkpoint, \textit{openai/clip-vit-base-patch32}, from Huggingface as well.

\section{Prompts in Use}
\label{sec:app-prompts-in-use}
In this section, we provide the exact prompts in use for baselines.

\subsection{Prompts for QA Conversion to Evidence Statement}
\label{sec:app-prompt-qa-to-evid}

\begin{figure*}[ht!] 
	\centering
	\includegraphics[width=\linewidth]{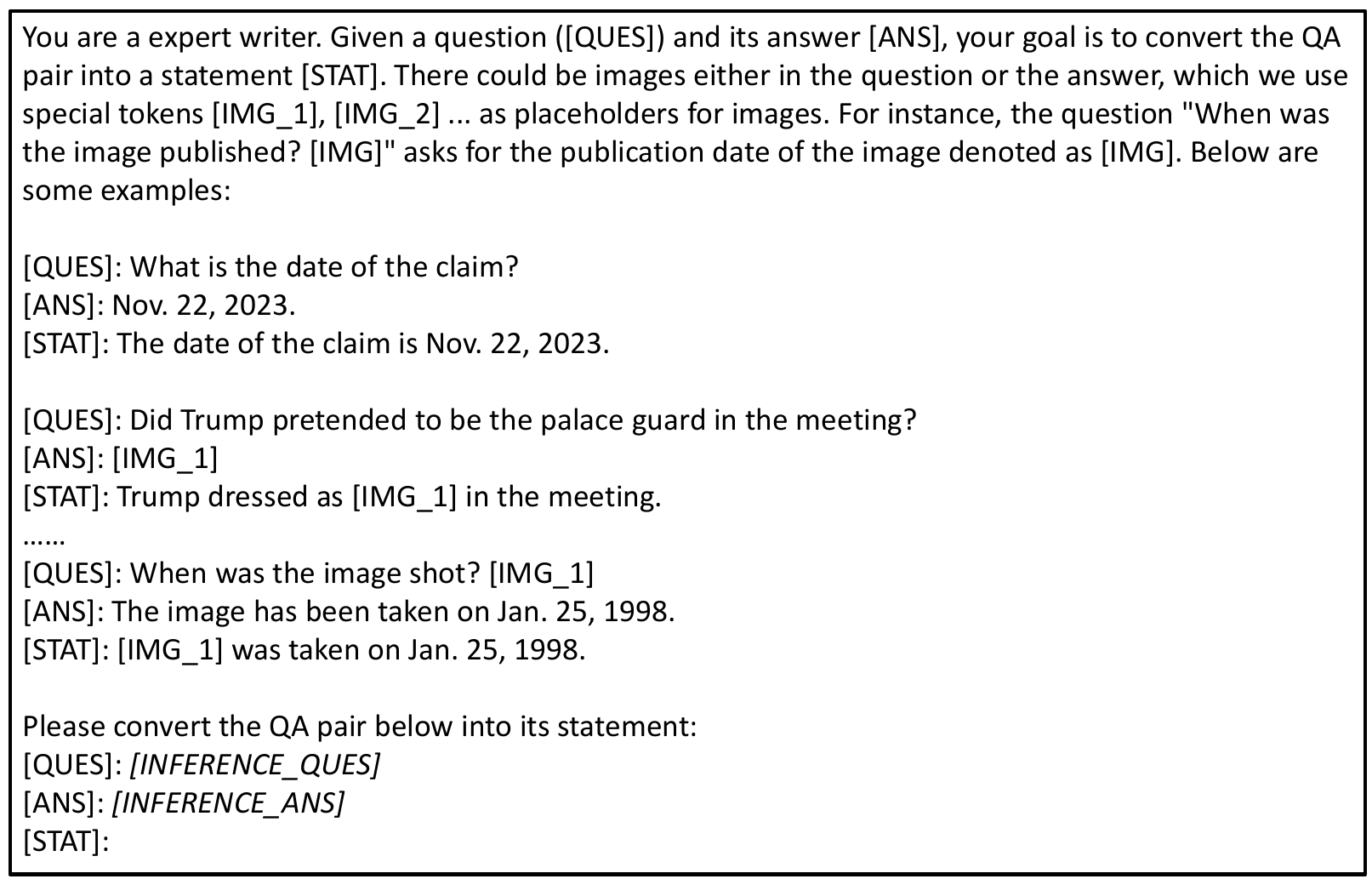} 
	\caption{
 \textbf{The prompt in use for converting QA pairs to evidence statement.}}
	\label{fig:appprompt-qa-to-evid}
\end{figure*}
Following~\citep{DBLP:journals/corr/abs-2411-05375}, we convert QA pairs to evidence statement, for both evidence evaluation and maintaining the evidence history. We consider a text-only conversion for simplicity and use special tokens as placeholders for image. These placeholders could be placed with the exact images in the future.
The prompt is demonstrated in Figure~\ref{fig:appprompt-qa-to-evid}.

\subsection{Prompts for Evaluation}
\label{sec:app-prompt-evaluation}
\begin{figure*}[ht!] 
	\centering
	\includegraphics[width=\linewidth]{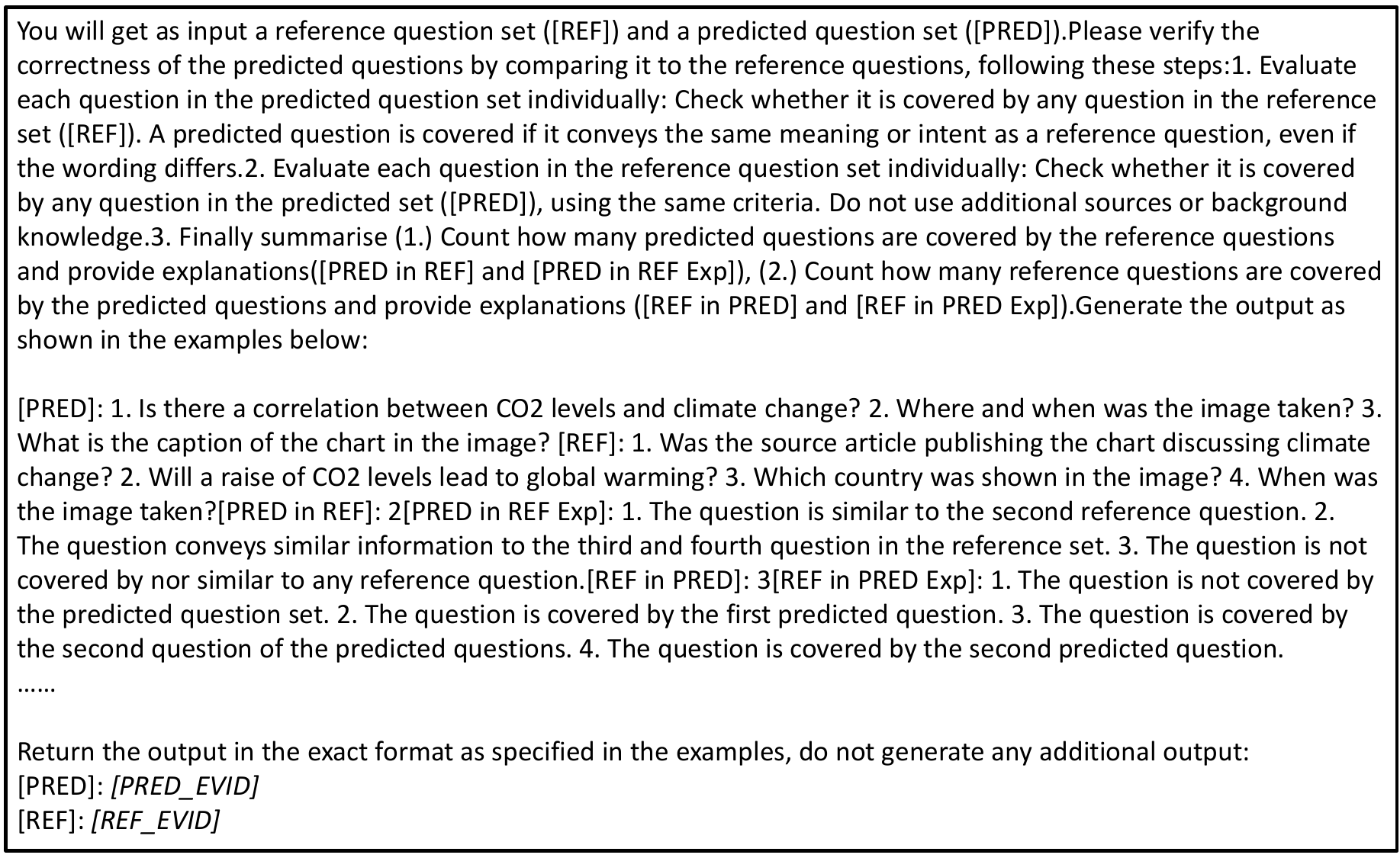} 
	\caption{
 \textbf{The evaluation prompt for generated questions.}}
	\label{fig:appprompt-eval-ques}
\end{figure*}
We adopt a reference based evaluation strategy, which compare predictions against references, for both question evaluation and evidence retrieval evaluation. 

\noindent\textbf{Question Evaluation. }Though questions could be multimodal, the semantics are the most informative. Therefore, we leverage a vanilla reference based evaluation scheme to compare the textual part of predicted questions and annotated questions. The exact prompt in use is shown in Figure~\ref{fig:appprompt-eval-ques}.

\begin{figure*}[ht!] 
	\centering
	\includegraphics[width=\linewidth]{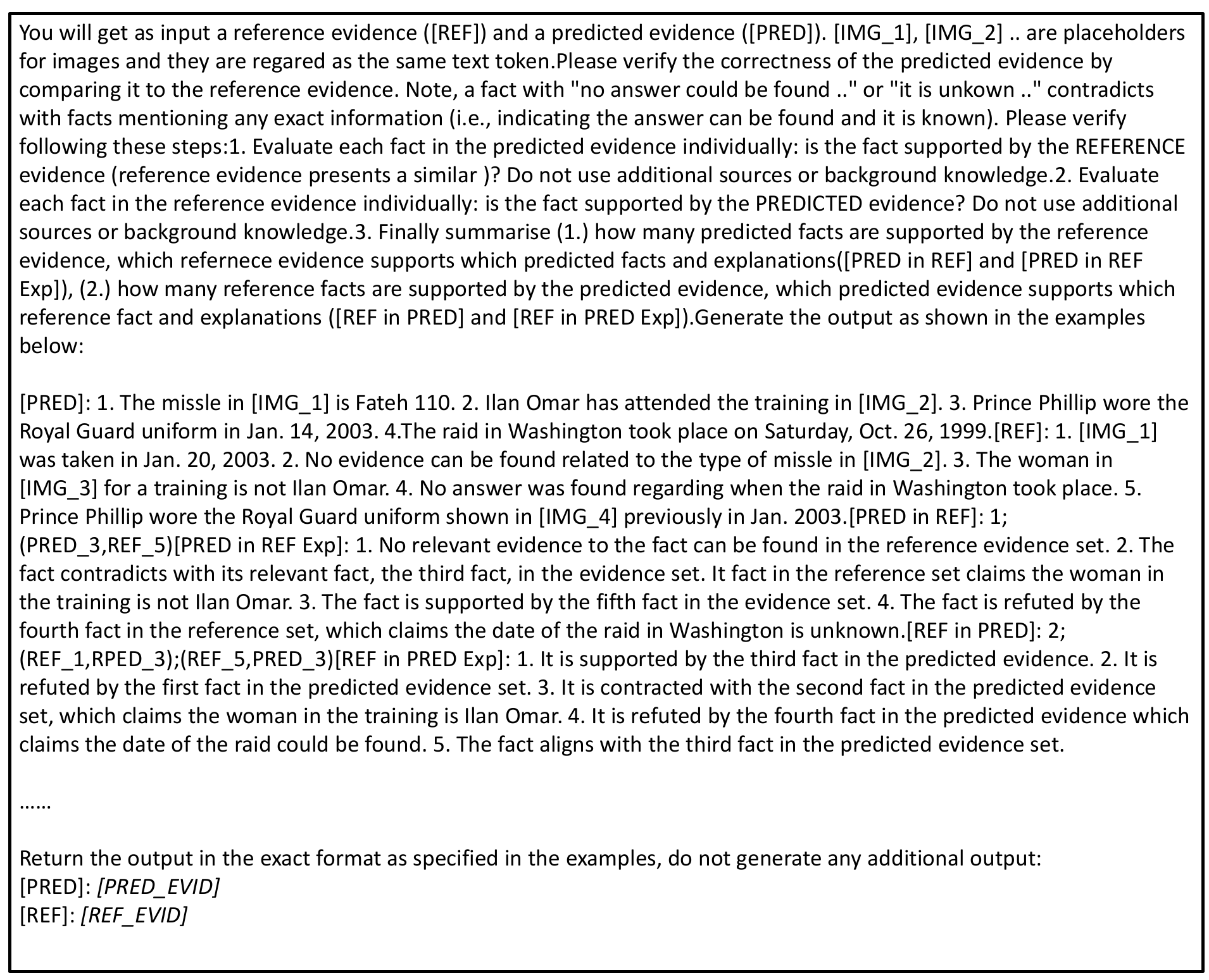} 
	\caption{
 \textbf{The evaluation prompt for retrieved evidence.}}
	\label{fig:appprompt-eval-evid}
\end{figure*}

\noindent\textbf{Evidence Evaluation. }We conducted a two-stage reference-based evaluation of evidence, as described in Section~\ref{sec:eval}.
In the first stage, we consider compare the textual of retrieved evidence and ground-truth evidence. Hence, the prompt used for evaluation is similar to that in question evaluation. The difference is that there are special image tokens in evidence and we need the evaluator to output the index of aligned predictions and ground-truth annotations. The prompt is illustrated in Figure~\ref{fig:appprompt-eval-evid}.

\subsection{Prompts for Question Generation}
\label{sec:app-prompt-qg}

\begin{figure*}[ht!] 
	\centering
	\includegraphics[width=\linewidth]{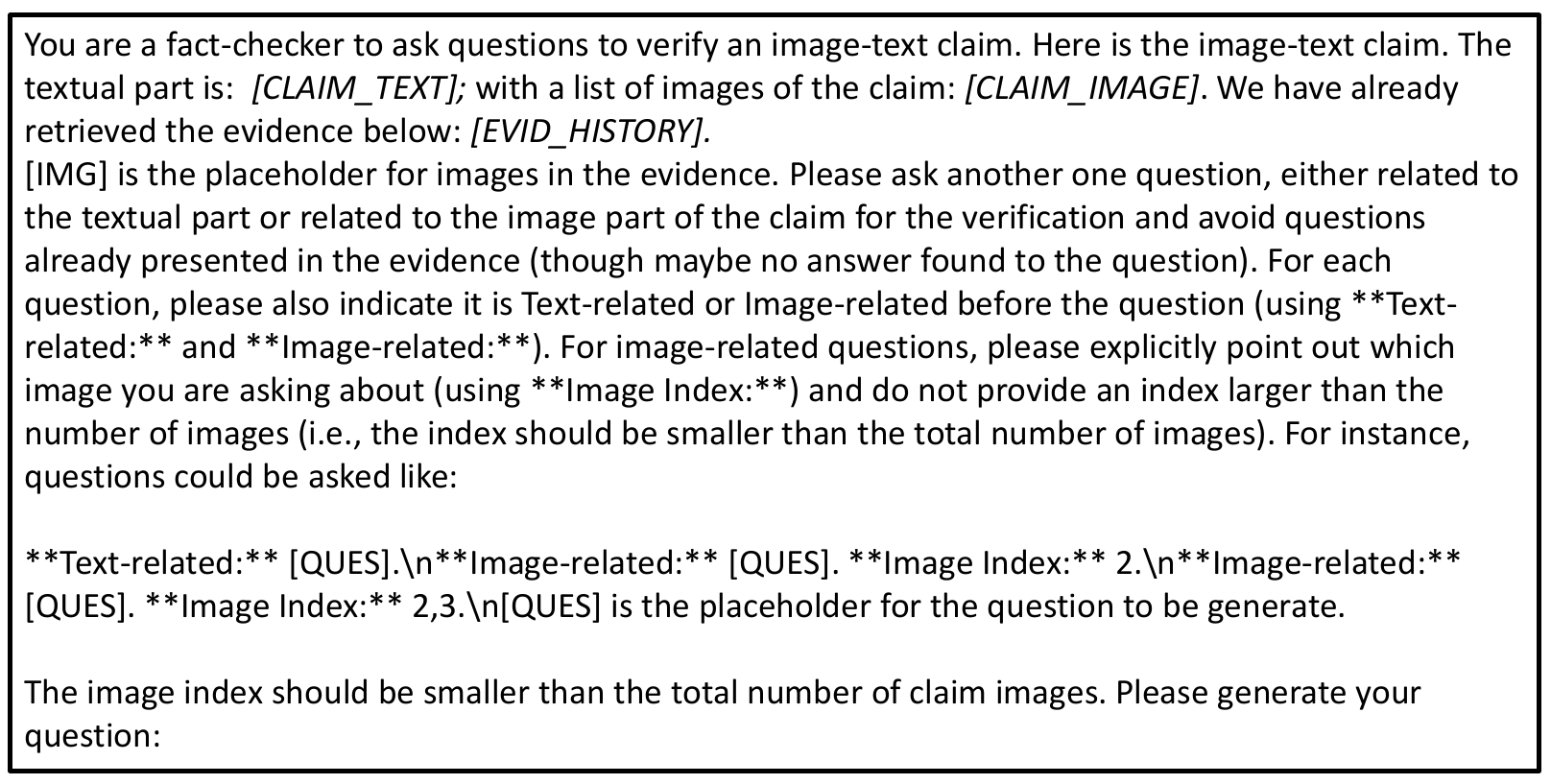} 
	\caption{
 \textbf{The prompt for dynamic question generation.} In the few-shot setting, the second paragraph is replaced with the few-shot demonstrations.}
	\label{fig:appprompt-qg-dynamic}
\end{figure*}

\begin{figure*}[ht!] 
	\centering
	\includegraphics[width=\linewidth]{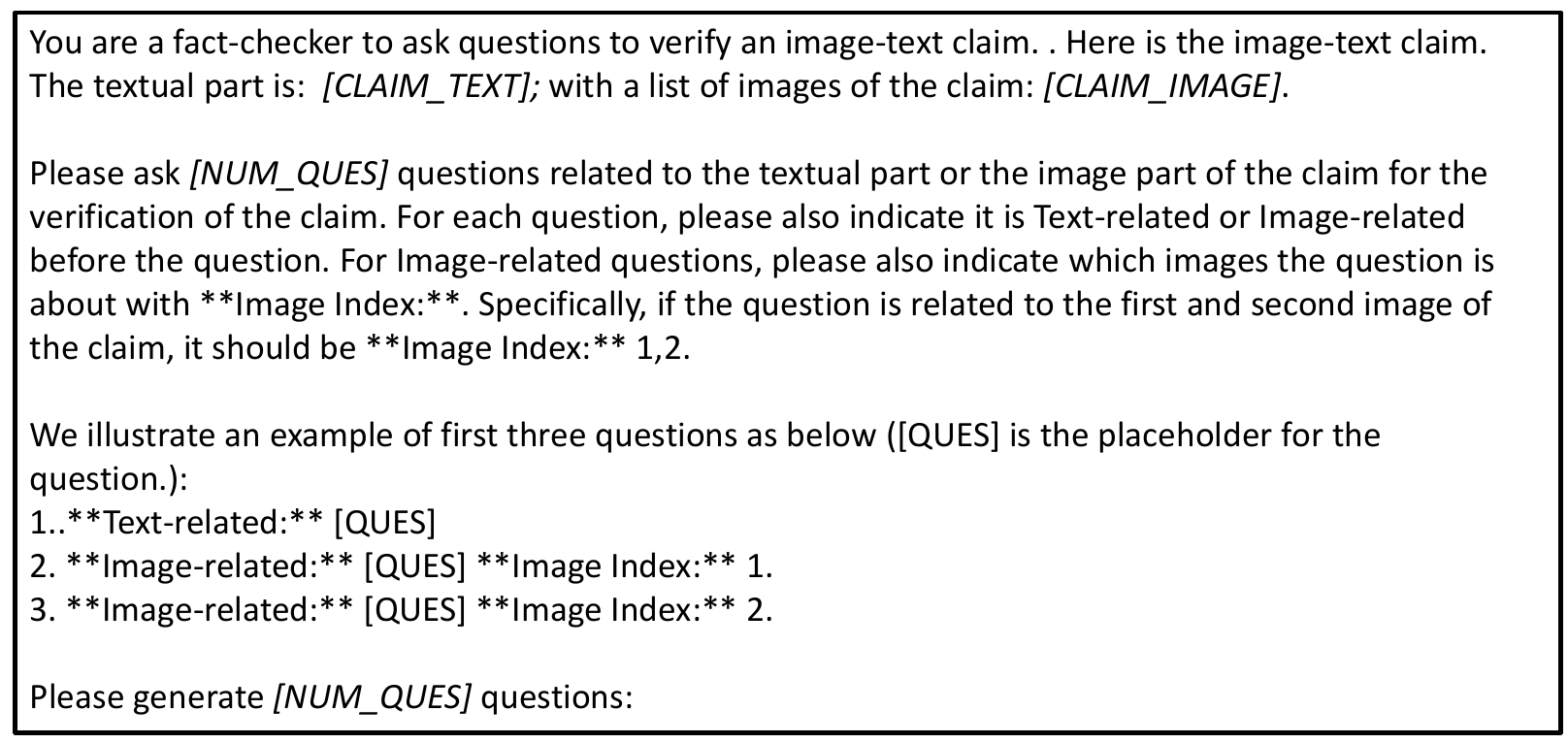} 
	\caption{
 \textbf{The prompt for paralleled question generation.} In the few-shot setting, the third paragraph is replaced with the few-shot demonstrations.}
	\label{fig:appprompt-qg-para}
\end{figure*}

We considered three strategies for question generation as introduced in Section~\ref{sec:exp-baseline}. The \textit{hybrid} generation is the combination of the \textit{paralleled} and the \textit{dynamic} question generation strategy. Below, we provide prompts for the DQG and PQG strategies.
For DQG, we use the prompt shown in Figure~\ref{fig:appprompt-qg-dynamic} and the PQG prompt is shown in Figure~\ref{fig:appprompt-qg-para}.

For the few-shot question generation setting, we utilize the same prompts while adding a few demonstrations before the information of inference instances.

\subsection{Prompts for Tool Selection}
\label{sec:app-prompt-tool-sel}
\begin{figure*}[ht!] 
	\centering
	\includegraphics[width=\linewidth]{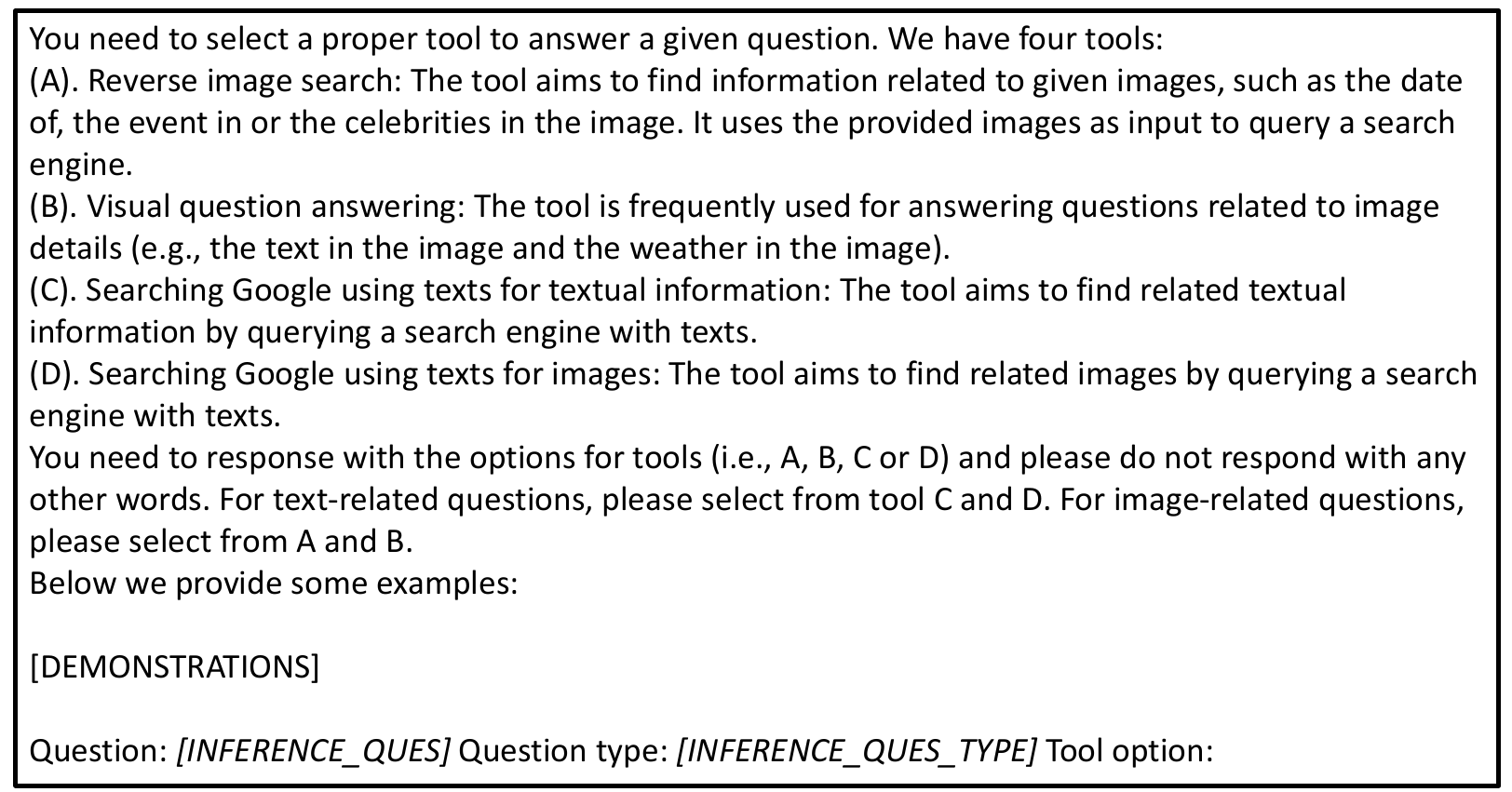} 
	\caption{
 \textbf{The prompt in use for selecting tools to answer questions.} [DEMONSTRATIONS] are placeholders of examples provided to guide the tool selection.}
	\label{fig:appprompt-tool-sel}
\end{figure*}
As mentioned in Section~\ref{sec:exp-results}, we observed bias of models for heavily relying on VQA as the answering tool, diverging from fact-checkers' choice. This leads to failures for retrieving essential evidence. 

Considering the issue, besides the tool definitions, we provide a few demonstrations, each consisting of a question, a question type and the tool should be selected. Specifically, we leverage the annotated metadata information of questions. For questions annotated with the \textit{answering method} of image-search, we consider the tool of RIS for such cases. For questions with the answering method as text-search while the answers are not images, we regard WST as the tool to be selected; if there are image answers, then the WSI should be the tool. For questions answered by image analysis, we would consider VQA as the answering method. 

The prompt for tool selection is shown in Figure~\ref{fig:appprompt-tool-sel}.
\subsection{Prompts for Answer Generation}
\label{sec:app-prompt-ans-gen}
As introduced in Section~\ref{sec:exp-baseline}, when leveraging the tools of RIS, WST and WSI, there follows an answering model (either an LLM or an MLLM) to leverage retrieved evidence to address the question.

For using an LLM to leverage textual evidence, we use the prompt: \textit{You need to answer a question according to a set of retrieved documents.}
\textit{Question:} \texttt{[QUES]};
\textit{Document:} \texttt{[RETRIEVED\_DOC]}. 
\textit{If the question is not answerable according to the provided document, please answer as: No answer can be found. Start you answer as: **ANSWER:** }

For VQA with an MLLM, we prompt models with the template below for an answer:
\textit{Question:} \texttt{[QUES]};
\textit{Related images to the question:} \texttt{[IMAGES]}. 

\subsection{Prompts for Verdict Prediction}
\label{sec:app-prompt-verdict-gen}
The verifier receives the claim and the retrieved evidence for predicting a veracity label of the claim. Below is the prompt exploited for the verifier: 
\textit{You need to select a verdict for a given Image-Text claim when provided a set of evidence. [IMG] is a placeholder for images. We provide four verdict labels and the definitions of them below: Supported: The claim is supported by the evidence presented. Refuted: The claim (either the text or the image part) is contradicted by the evidence presented. Not Enough Evidence: There is not enough evidence (NEE) to support or refute the claim. Conflicting: The claim is misleading due to conflicting evidence/cherry-picking, but not explicitly refuted. You need to response with the verdict for the claim (i.e., Supported, Refuted, Not Enough Evidence or Conflicting) and please do not respond with any other words.}
\textit{The metadata of the claim:} \texttt{[DATE\_AND\_LOCATION]}.
\textit{Claim:} \texttt{[CLAIM\_TEXT]};
\textit{Claim images:} \texttt{[CLAIM\_IMAGES]}. 
\textit{Here is the evidence:} \texttt{[EVID]}. 
\textit{Verdict:}

\subsection{Prompts for Justification Generation}
\label{sec:app-prompt-justi-gen}
The prompt for justification generation receives the information about the claim (textual part and claim images), retrieved evidence and predicted verdict to explain how the verdict could be reached.
Below is the exact prompt in use:

\textit{Given an image-text claim and a set of evidence for verifying the claim, a fact-checker predict a veracity label for the claim. You need to explain how the verdict is reached for the image-text claim. Below is information for the image-text claim:} 
\textit{The metadata of the claim:} \texttt{[DATE\_AND\_LOCATION]}.
\textit{Claim:} \texttt{[CLAIM\_TEXT]};
\textit{Claim images:} \texttt{[CLAIM\_IMAGES]}. 
\textit{The predicted verdict is:} \texttt{[PRED\_VERDICT]}.
\textit{Here is the evidence:} \texttt{[EVID]}. \textit{Please generate your justification (i.e., explanation) for the verdict:}

Outputs from MLLMs are verbose, whereas human annotated justifications are concise. Therefore, we conduct one step further to prompt the corresponding LLMs to summarize the generated jsutifications in one or two sentences.

\end{document}